\RequirePackage{tipa}
\documentclass[sigconf]{acmart}
\usepackage{booktabs} 

\usepackage[utf8]{inputenc}
\usepackage[T1]{fontenc}
\usepackage{amsfonts}
\usepackage{amsmath}
\usepackage{amssymb}
\usepackage{booktabs}
\usepackage{color}
\usepackage{comment}
\usepackage{graphicx}
\usepackage{lipsum}
\usepackage{makecell}
\usepackage{microtype}
\usepackage{multirow}
\usepackage{nicefrac}
\usepackage{siunitx}
\usepackage{subfig}
\usepackage{url}
\usepackage{wrapfig}
\usepackage{xspace}
\usepackage{changepage}
\usepackage{tikz}
\usepackage{dblfloatfix}
\usepackage{pdflscape}
\usepackage{upgreek}
\usepackage{hyperref}
\usepackage{cleveref}

\makeatletter

\newcommand{\bx}{\mathbf{x}}
\newcommand{\by}{\mathbf{y}}

\newcommand{\D}{D_{\mathcal{Y}}}
\newcommand{\Ex}{\mathop{\mathbb{E}}}

\newcommand{\dictentry}[4]{\textbf{#1}\markboth{#1}{#1}\ {(#2)}\ \textit{#3}\ $\bullet$\ {#4}}

\definecolor{mcolor}{rgb}{0.86,0.71,0.51}
\definecolor{mcolor}{rgb}{0.86,0.71,0.51}
\definecolor{mcolor}{rgb}{0.86,0.71,0.51}

\definecolor{mcolor}{rgb}{0.86,0.71,0.51}
\newcommand\textim[2][fill=mcolor!50]{%
    \tikz[baseline]\node[%
        inner ysep=0pt,
        inner xsep=8pt,
        anchor=text,
        rectangle,
        rounded corners=1mm,
        #1] {\strut#2};%
}

\newcommand{\parag}[1]{\noindent\textbf{#1}\quad}

\DeclareRobustCommand\onedot{\futurelet\@let@token\@onedot}
\def\@onedot{\ifx\@let@token.\else.\null\fi\xspace}

\def\eg{\emph{e.g}\onedot} 
\def\ie{\emph{i.e}\onedot}

\def\etal{\emph{et al}\onedot}
\def\viz{\emph{viz}\onedot}

\makeatother

\acmConference[ICVGIP 2018]{11th Indian Conference on Computer Vision, Graphics and Image Processing}{December 18--22, 2018}{Hyderabad, India}
\acmBooktitle{11th Indian Conference on Computer Vision, Graphics and Image Processing (ICVGIP 2018), December 18--22, 2018, Hyderabad, India}
\copyrightyear{2018}
\acmYear{2018}
\setcopyright{acmlicensed}
\acmPrice{15.00}
\acmDOI{10.1145/3293353.3293386}
\acmISBN{978-1-4503-6615-1/18/12}
\acmArticle{33}

\editor{Anoop M. Namboodiri}
\editor{Vineeth Balasubramanian}
\editor{Amit Roy-Chowdhury}
\editor{Guido Gerig}

\begin{document}

\author{Ankush Gupta}
\affiliation{%
  \department{Visual Geometry Group}
  \institution{University of Oxford}}
\email{ankush@robots.ox.ac.uk}

\author{Andrea Vedaldi}
\affiliation{%
  \department{Visual Geometry Group}
  \institution{University of Oxford}}
\email{vedaldi@robots.ox.ac.uk}

\author{Andrew Zisserman}
\affiliation{%
  \department{Visual Geometry Group}
  \institution{University of Oxford}}
\email{az@robots.ox.ac.uk}

\renewcommand{\shortauthors}{Gupta, Vedaldi, Zisserman}

\begin{abstract}
This work presents a method for visual text recognition without
using any paired supervisory data.
We formulate the text recognition task as one of aligning the
conditional distribution of strings predicted from given text images,
with lexically valid strings sampled from target corpora.
This enables fully automated, and unsupervised learning from just line-level text-images, and unpaired text-string samples, obviating the need for large aligned datasets.
We present detailed analysis for various aspects of the proposed method,
namely --- (1) impact of the length of training sequences on convergence,
(2) relation between character frequencies and the order in which they are learnt,
(3) generalisation ability of our recognition network to inputs of arbitrary lengths,
and (4) impact of varying the text corpus on recognition accuracy.
Finally, we demonstrate excellent text recognition accuracy on both
synthetically generated text images, and scanned images of real printed books,
using no labelled training examples.
\end{abstract}



\title[Learning to Read by Spelling]{Learning to Read by Spelling}
\subtitle{Towards Unsupervised Text Recognition}

\begin{teaserfigure}
  \centering
  \includegraphics[width=0.8\textwidth]{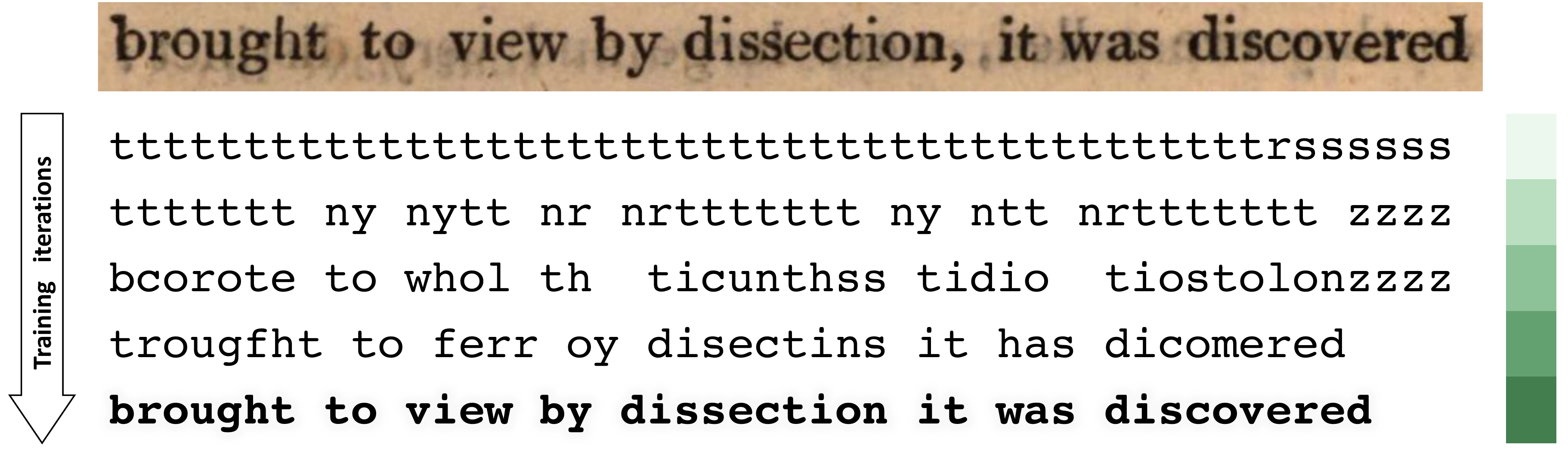}
  \caption{Text recognition from unaligned data. \textmd{We present a method for
  recognising text in images without using any labelled data. This is achieved
  by learning to align the statistics of the predicted text strings, against
  the statistics of valid text strings sampled from a corpus.
  The figure above visualises the transcriptions
  as various characters are learnt through the training iterations.
  The model first learns the concept of \texttt{\{space\}},
  and hence, learns to segment the string into words; followed by
  common words like \texttt{\{to, it\}}, and only later learns
  to correctly map the less frequent characters like \texttt{\{v, w\}}.
  The last transcription also corresponds to the ground-truth (punctuations
  are not modelled).
  The colour bar on the right indicates the accuracy (darker means higher accuracy).}}
  \vspace{1em}
\end{teaserfigure}

\maketitle

\section{Introduction}
\begin{adjustwidth}{8mm}{}
\dictentry{read}{ri\textlengthmark d}{verb}{Look at and comprehend the meaning
of (written or printed matter) by interpreting the characters or symbols of which it is composed.}\\
\dictentry{spell}{sp\textepsilon l}{verb}{Write or name the letters that form (a word) in correct sequence.}\\
\null\hfill --- \emph{Oxford Dictionary of English}
\end{adjustwidth}

\noindent Text recognition, namely the problem of reading text in images,
is a classic problem in pattern recognition and computer vision that has enjoyed
continued interest over the years, owing to its many practical applications,
such as recognising printed~\cite{tesseract,Smith07} or handwritten~\cite{Lecun89,bunke04}
documents, or more recently, text in natural images~\cite{Jaderberg16,Mishra12,Neumann12}.
Consequently, many different and increasingly accurate methods have been developed.
Yet, all such methods adopt the same \emph{supervised learning} approach that requires
example images of text annotated with the corresponding strings.

Annotations are expensive because they must be \emph{aligned} to individual training images.
For example, for a text-image of \textim{\textbf{\emph{cats}}}, the corresponding
annotation is the string \texttt{\{c,a,t,s\}}.
A straightforward but tedious approach is to collect such annotations
manually~\cite{Karatzas13,netzer2011reading,Wang10b};
however, since datasets often comprise several million
examples~\cite{Krizhevsky12,Jaderberg14c}, this scales poorly.
Another, perhaps more pragmatic, approach is to engineer highly-sophisticated
synthetic data generators to mimic real images~\cite{Gupta16,Jaderberg14c,Wang12}.
However, this requires developing new generators for each new textual domain,
and could be problematic for special cases such as text in ancient manuscripts.

\begin{figure}[t]
  \centering
  \includegraphics[width=\linewidth]{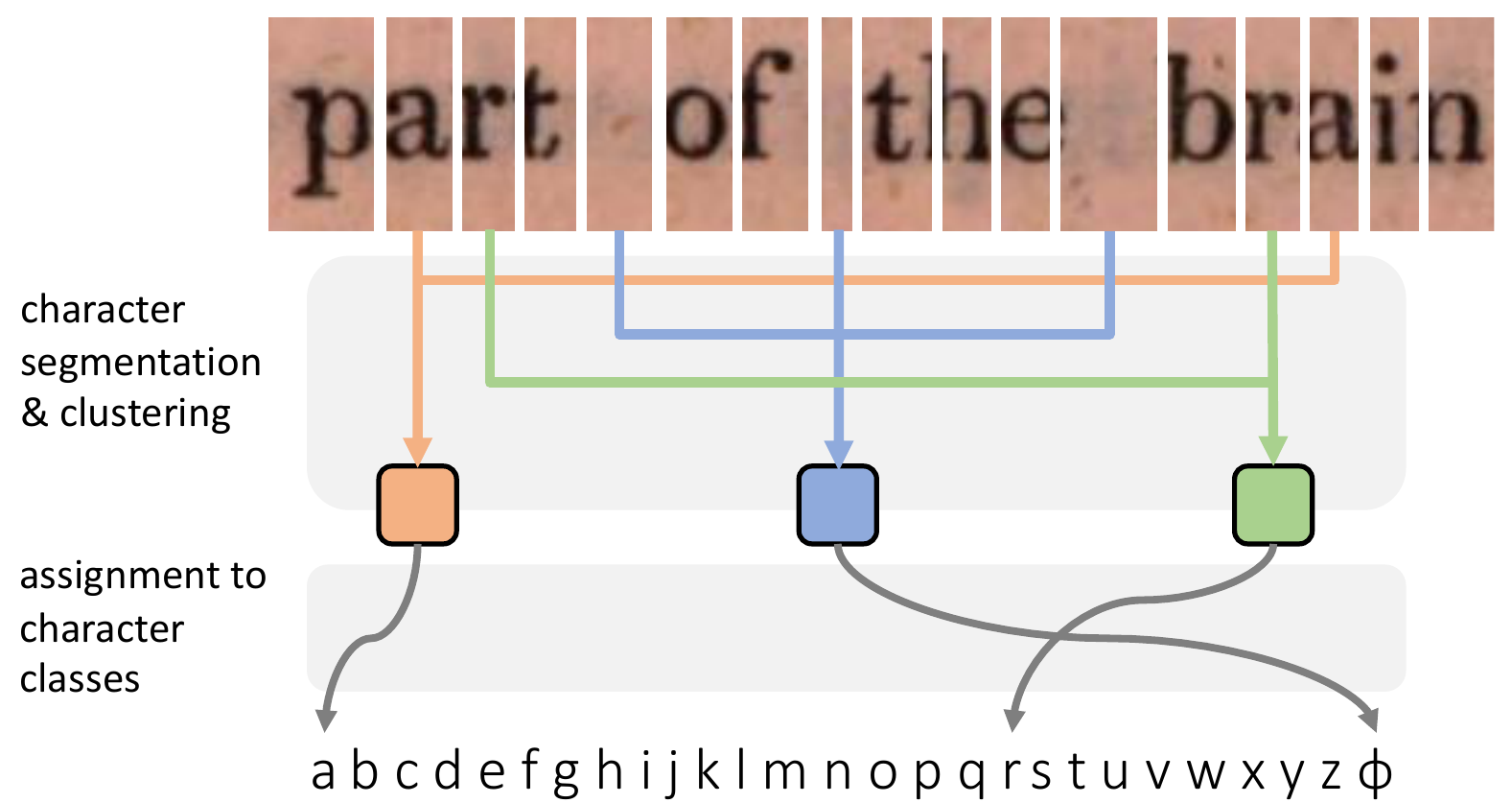}
  \vspace{-5mm}
  \caption{Unsupervised text recognition \textmd{can be factored into two sub-problems:
  (1)~\emph{visual}:~segmentation at the character-level, followed by
  clustering (or recognition) into a known number of classes, and
  (2)~\emph{linguistic}:~determining the character identity of these clusters
  based on language constraints. Three character classes corresponding to
  \texttt{\{a,$\phi$,r\}} are visualised above ($\phi$ stands for \text{\{space\}}).
  A given text image is mapped to a sequence of characters using a fully-convolutional network;
  the predicted sequences are compared against linguistically valid text-strings
  using an \emph{adversarial discriminator}, which guides the mapping of
  the characters to the correct identity.
  The two networks trained end-to-end jointly, enable text recognition
  without any labelled training data.}}
  \label{fig:method}
  \vspace{-5mm}
\end{figure}

\begin{figure*}[t]
  \centering
  \includegraphics[width=\textwidth]{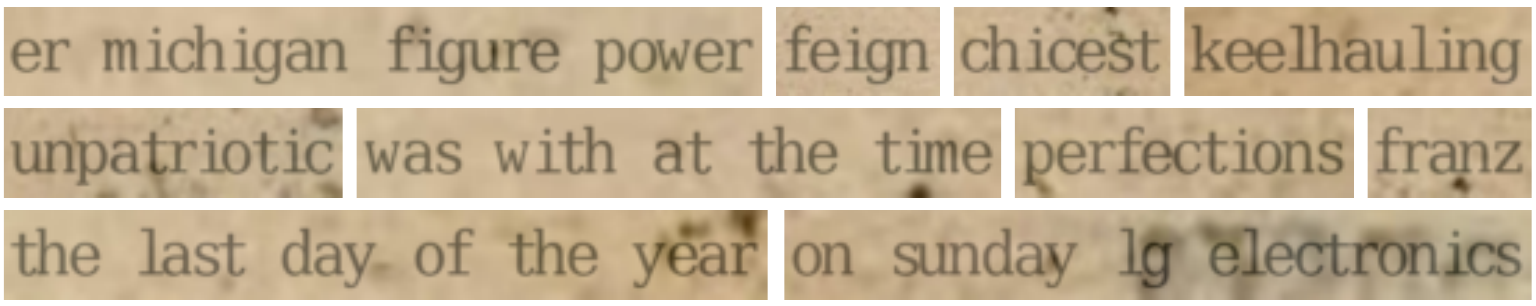}
  \vspace{-5mm}
  \caption{Synthetic text-image samples.
    \textmd{A few synthetically generated samples of different lengths,
     used in the controlled experiments (see~\cref{sec:exp}).
     Our model attains ${\approx} 99\%$ \emph{character accuracy} and ${\approx} 95\%$
     \emph{word accuracy} on such samples (\cref{sec:wlen-generalise}),
     after training on only \emph{unaligned} image and text examples.}}
  \label{fig:synth}
\end{figure*}

We propose instead to develop learning algorithms that can work with \emph{unaligned}
annotations. In this paradigm, images containing text can be extracted \eg from
scanned documents or by mining online image collections~\cite{Jaderberg14}.
Independently, strings containing the same \emph{type} of text (but not exactly
the same text) can be readily harvested from machine readable text corpora~(\eg WMT datasets~\cite{wmt}).
Both steps can be implemented economically in a fully-automated manner,
making such an approach highly desirable.

More specifically, we demonstrate visual text recognition by only providing
examples of valid textual strings, but without requiring them to be aligned to
the example images. In this manner, the method is almost unsupervised, as by only
knowing how to \textbf{spell} correctly, it learns to \textbf{read}.
The method works by learning a predictor that converts images into strings that
\emph{statistically} match the target corpora, implicitly reproducing quantities
such as letter and word frequencies, and n-grams. We show empirically that this seemingly
weak principle is in fact sufficient to drive learning successfully (\cref{sec:exp}).

Text recognition can be factored into two sub-problems (see~\cref{fig:method}):
(1) \emph{visual}:~segmenting the text-image into characters and clustering
the different characters into a known number of distinct classes,
and (2) \emph{linguistic}:~assigning these clusters to the correct character identity.
Indeed, earlier attempts at unsupervised text recognition proposed two-stage solutions
corresponding to the two sub-problems~\cite{Huang07c,Kae09,Knight11,Aldarrab17}.
We address the first problem by exploiting the properties of standard fully-convolutional
networks~\cite{Long15} --- namely locality and translation invariance of the network's filters.
The second problem is equivalent to solving for the correct permutation,
or breaking a 1:1--substitution cipher~\cite{Peleg79}.
The latter problem is NP-hard under a bi-gram language model~\cite{Nuhn13}.
While several solutions like aligning uni-gram (\ie frequency matching) or n-gram
statistics~\cite{sutskever16,liu17} have been proposed traditionally for breaking
ciphers~\cite{dooley2013brief}, we instead adopt an adversarial approach~\cite{goodfellow2014generative}.
The result is a compact fully-convolutional sequence (\ie multiple words/text-string)
recognition network which is trained against a discriminator in an end-to-end fashion.
The discriminator uses as input only unaligned examples of valid text strings.

We study various factors which affect training convergence, and use
synthetically-generated data for these controlled experiments.
We also show excellent recognition performance on real text images from the
Google1000 dataset~\cite{Google1k}, given \emph{no aligned labelled data}.

The rest of the paper is structured as follows.
\Cref{sec:priorart} reviews related work, \cref{sec:method} describes our
technical approach, \cref{sec:impl} gives the implementation details, \cref{sec:exp} evaluates the method on the aforementioned
data, and \cref{sec:conclude} summarises our findings.

\section{Related Work}\label{sec:priorart}

\parag{Supervised Text Recognition.} Distinct paradigms have emerg- ed and
evolved in text recognition. Traditional character-level methods adopt either
sliding-window classifiers~\cite{Wang11,Mishra12,Yao14,Jaderberg14,Wang12},
or over / under segment into parts~\cite{Bissacco13,Alsharif14,Neumann12},
followed by grouping through classification. Words or sentences are then
inferred using language models~\cite{Wang11,Wang10b,Lee14,
Alsharif14,Jaderberg14,Wang12,Mishra12,Mishra12a,Shi13,Novikova12}. Another
set of methods process a whole word image, modelling it either as retrieval in a
collection of word images from a fixed lexicon~\cite{Goel13,Almazan14,Rodriguez15, Gordo15}
or as learning multiple position dependent classifiers~\cite{Jaderberg14c,Jaderberg15a,Poznanski16}.
Our recognition model is similar to these character-sequence classifiers in that
we train with a fixed number of output characters; but there is an important
difference: we discard their fully-connected (hence, position sensitive)
classifier layers and replace them with fully-convolutional layers.
This drastically reduces the number of model parameters, and lends generalisation
ability to inputs of arbitrary length during inference. More recent methods treat
the text-recognition problem as one of sequence prediction in an encoder-decoder
framework~\cite{Cho14,Sutskever14}. \cite{Su14} adopted this framework first,
using HOG features with Connectionist Temporal Classification (CTC)~\cite{Graves06}
to align the predicted characters with the image features.
\cite{He16DTRN,ShiBY15} replaced HOG features with stronger CNN features,
while~\cite{Lee16,Shi16} have adopted the soft-attention~\cite{Bahdanau15} based
recurrent decoders. Note, all these methods learn from labelled training examples.\\

\parag{Unsupervised Text Recognition.}
Unsupervised methods for text recognition can be classified into two categories.
First, category includes generative models for document images. A prime example
is the \emph{Ocular} system~\cite{berg2013unsupervised}, which jointly models the text content,
as well as the noisy rendering process for historical documents, and infers the
parameters through the EM-algorithm~\cite{Dempster77}, aided by an n-gram language model.
The second category includes methods for automatic decipherment.
Decipherment, is the process of mapping unintelligible symbols
(ciphertext) to known alphabet/language (plaintext).
When the input is visual symbols, it becomes equivalent to text recognition.
Some early works~\cite{Nagy86,Casey86} for optical character recognition (OCR),
indeed model it as such. \cite{Ho00} cluster connected components in binarised
document images and assign them to characters by maximising overlap with
a fixed lexicon of words based on character frequencies and co-occurrence;
\cite{Huang07c,Kae09} also follow the same general approach.
\cite{Aldarrab17} break the Borg cipher, a 17th century 408-pages manuscript,
by also first clustering symbols but decipher using
the noisy-channel framework of~\cite{Knight06} through finite-state-machines.
\cite{Lee02} learn mappings from hidden-states
of an HMM with their transition probabilities initialised with
conditional bi-gram distributions.
\cite{Kozielski14} propose an iterative scheme for
bootstrapping predictions for learning HMMs models, and recognise handwritten text.
However, their approach is limited to (1) word images,
(2) fixed lexicon ($\approx$44K words) to facilitate exhaustive tree search,
whereas, our method is applicable to full \emph{text strings},
does not require a pre-defined lexicon of words.\\

\parag{Unsupervised Learning by Matching Distributions.}
Output Distribution Matching (ODM) which aligns the \emph{distributions}
of predictions with \emph{distributions} of labels was proposed in~\cite{sutskever16}
for ``principled'' unsupervised learning; although similar ideas for learning by
matching statistics have been explored earlier, \eg for decipherment (see above),
and also for machine translation~\cite{Snyder10,Ravi08}.
\cite{liu17} extend ODM to sequences, and apply it to OCR with known character
segmentations and pre-trained image features.
In essence, ODM~\cite{sutskever16}, or Empirical-ODM~\cite{liu17} minimises the
KL-divergence cost between the empirical predicted and ground-truth
n-gram distributions.
Our learning principle is the same, however, we do not explicitly
formulate the matching cost, instead learn it online using an adversary~\cite{goodfellow2014generative}.
Recent works~\cite{Artetxe17,Lample17} have demonstrated
unsupervised machine translation using such adversarial losses, however they
closely follow the \emph{CycleGAN} framework~\cite{Zhu17}
which learns a bidirectional mapping between the input and target
domains to enforce bijection. This framework has also been applied recently in
\emph{CipherGAN} to break ciphers~\cite{Gomez18}.
The \emph{CycleGAN} framework learns a bi-directional mapping to enforce strong
correlation between the input and the generated output to avoid the
degenerate failure mode of collapsing to the same output instance regardless of
the input. We, however, dispense with back-translation/reconstruction, and
instead enforce correlation directly in the structure of the recogniser by
limiting the receptive-field of convolutional layers.
Hence, our method is an instantiation of the original (single)
generator--discriminator framework of \emph{GANs}~\cite{goodfellow2014generative}.
However, our method is perhaps the first to decode sequences of discrete symbols
from images using an adversarial framework; these two domains have only been
explored independently in \emph{CycleGAN} and \emph{CipherGAN} respectively.

\section{Method}\label{sec:method}
The aim of text recognition is to predict a sequence of characters given an image
of text. Let the image be a tensor $\bx \in \mathcal{X} = \mathbb{R}^{H\times W\times C}$,
where $H$, $W$, $C$ are its height, width, and number of colour channel(s) respectively.
Furthermore, let $\by = (y_1, y_2, \hdots, y_n) \in \mathcal{Y}$ denote the
corresponding character string where each $y_i$ is a character from an alphabet
$\mathcal{A}$ containing $K$ symbols, \ie $|\mathcal{A}| = K$.
For later convenience, a character $y_i$ is represented as a $K$-dimensional
one-hot vector. Since such vectors are elements of the $K$-dimensional simplex
$\Delta^K$, we set $\mathcal{Y} =(\Delta^K)^n \subset \mathbb{R}^{K\times n}$.
Without loss of generality, we consider strings of a fixed length $n \in \mathbb{N}$.
The objective of \emph{unsupervised} text recognition, then, is to learn the
mapping $\Phi(\bx) = \by$, given only \emph{unpaired} examples from the two
domains $\{\bx_i\}_{i=1}^N$ where $\bx_i \in \mathcal{X}$ and $\{\by_j\}_{j=1}^M$
where $\by_j \in \mathcal{Y}$.

We cast this in an adversarial learning framework based on Goodfellow~\etal~\cite{goodfellow2014generative}.
We view the \emph{text recogniser} $\Phi: \mathcal{X} \rightarrow \mathcal{Y}$
as a \emph{conditional generator} of strings $\by$.
The recogniser competes against an adversarial discriminator $\D$, which aims to
distinguish between \emph{real} strings $\{\by\}$ and \emph{generated} strings $\{\Phi(\bx)\}$.
In other words, $\Phi$ and $\D$ are optimised simultaneously to play the following
two-player minimax game~\cite{goodfellow2014generative}
$\min_{\Phi}\max_{\D} \mathcal{L}(\Phi, \D)$ where the value function is given by:
\begin{equation*}\label{eq:gan}
  \mathcal{L}(\Phi, \D) = \Ex_{\by\thicksim \mathcal{Y}}[\log\D(\by)] + \Ex_{\bx \thicksim \mathcal{X}}[\log(1 - \D(\Phi(\bx))].
\end{equation*}

The recogniser learns the visual problem of segmenting characters in
images, and organising them into distinct categories;
while the discriminator, by checking the predicted sequence of characters
against linguistically valid strings,
guides the assignment of these categories into the respective
correct character classes (see \cref{fig:method}).\\

\begin{figure}[t]
  \centering
  \includegraphics[width=\linewidth]{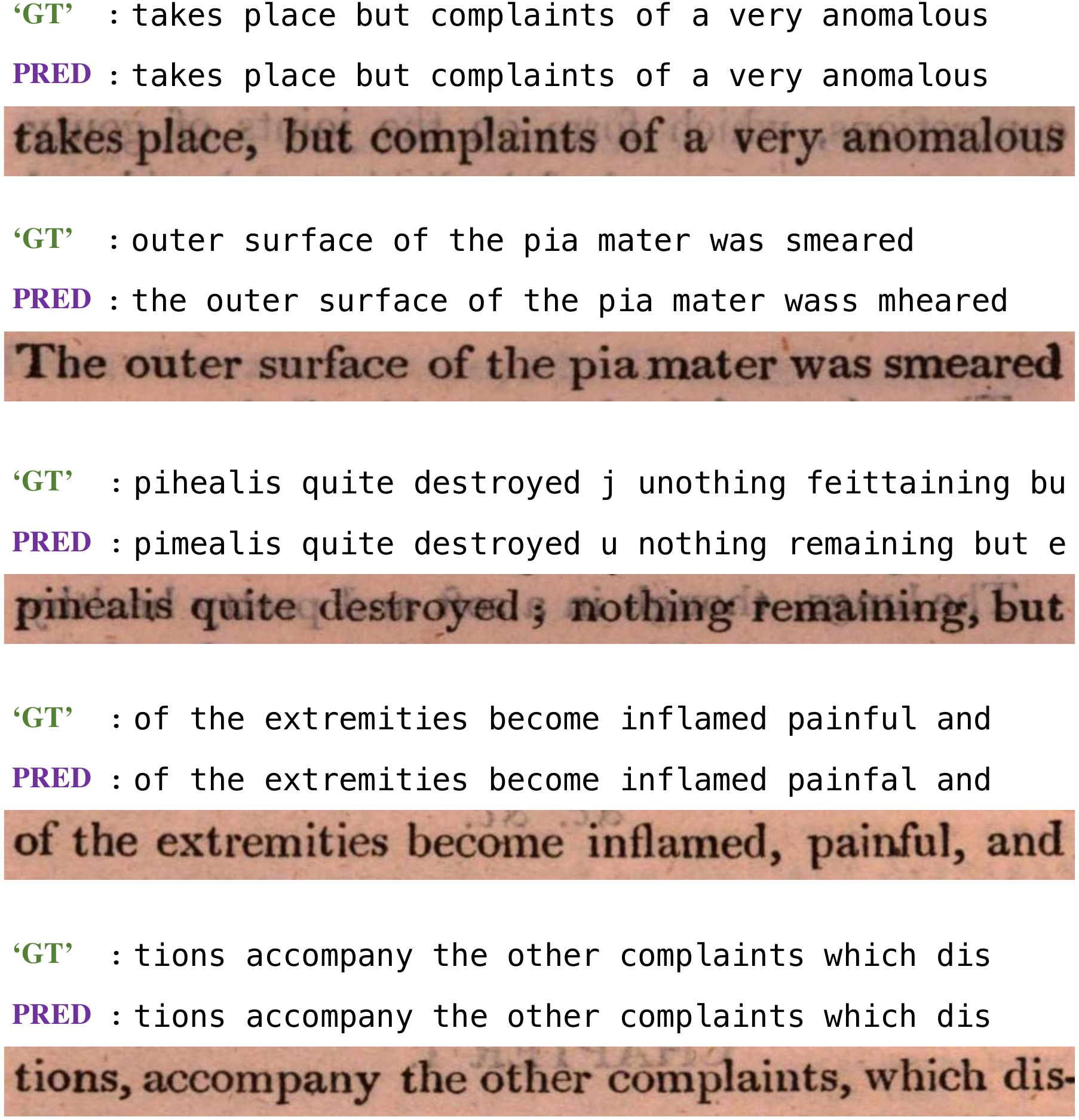}
  \vspace{-5mm}
  \caption{Real text-image samples.
  \textmd{Randomly selected samples from a \emph{real} scanned book's test set
    along with the ``ground-truth'' (\texttt{`GT'}) and the predicted strings (\texttt{PRED});
    punctuations are not modelled.
    Our model achieves excellent recognition performance ---
    $96.2\%$ \emph{character}, and $84.8\%$ \emph{word} accuracy (see~\cref{sec:real}) without
    using any aligned/labelled training examples.
    Note the ``ground-truth'' (\texttt{`GT'}) comes from Google's OCR engine output,
    hence is not perfect (\eg second and third image above).}}
  \label{fig:real}
  \vspace{-6mm}
\end{figure}

\parag{Grounding.} A potential pitfall is that the string generator network
(or recogniser $\Phi$) may learn to use the input image as a mere source of noise,
using it to generate the correct distribution of strings, without learning to
recognise the string represented in the image.
A useful mapping, instead, must be \emph{grounded}, \ie the generated string $\by = \Phi(\bx)$
should correspond to the text represented in the input image $\bx$.

A possible way to encourage grounding is to ensure that the image $\bx$ can be recovered
back from the string $\by$.
Both \emph{CycleGAN}~\cite{Zhu17} and \emph{CipherGAN}~\cite{Gomez18} achieve this by
learning a second inverse mapping $\Psi: \mathcal{Y} \rightarrow \mathcal{X}$
from the target domain back to the input and complete the cycle $\Psi(\Phi(\bx)) \triangleq \bx$.
However, learning a mapping from character strings to images is highly ambiguous:
rendering a given string requires sampling the background image, font style,
font colour, geometry of the glyphs, shadows, noise etc.
This ambiguity arises because text recognition requires translating between
two very different \emph{modalities}, \viz text and images,
which is much harder than translating within the same modality,
\eg between images in \emph{CycleGAN}~\cite{Zhu17} where only local texture is
modified, or between character strings in \emph{CipherGAN}~\cite{Gomez18},
where the characters are permuted.

Instead of enforcing cycle-consistency, we encourage grounding via
the following two key architectural modifications in the recogniser $\Phi$ (architectural details are given in~\cref{sec:impl}):

\begin{enumerate}
  \item \textbf{Prediction Locality.} The character predictor is local, with a
  receptive field large enough to contain at most two or three characters in the image.
  While this may sound simple, it embodies a powerful constraint.
  Namely, such local predictors can generate a string which is globally
  consistent only if they correctly transduce the structure of the underlying image.
  Otherwise, local predictors may be able to match local text statistics such as
  $n$-grams, but would not be able to match  global text statistics, such as
  forming proper words and sentences~(see also section~6.1 of~\cite{sutskever16} for similar ideas).
  Global consistency is enforced by the adversarial discrminiator which has a large
  receptive field over the predicted characters (see~\cref{sec:impl}).

  \item \textbf{Reduced Stochasticity.}
  We also make the generated strings a deterministic function of the input.
  We achieve this by removing the noise input from $\Phi$ which is normally used in generator networks.
  Furthermore,  we do not use dropout regularization~\cite{Srivastava15}.
\end{enumerate}

\parag{Training Objective.} The discriminator $\D$ operates in the domain of
\emph{discrete} symbols. While the real symbols are represented as one-hot vectors
or \emph{vertices} $\operatorname{Vert}(\Delta^K)$ of the standard simplex, the
generated symbols are output of a $\operatorname{SoftMax}$ operator over predicted
logits, and hence typically belong to the \emph{interior} of the simplex $\Delta^K$.
This was identified, as the cause for \emph{uninformative discrimination} in
\emph{CipherGAN}~\cite{Gomez18}, where the discriminator distinguishes using this
unimportant difference, rather than soundness of the generated strings.

To mitigate this, we adopt their proposed solution and learn a $d$-dimensional
embedding for each of the $K$ symbols in the alphabet, collectively represented
by a matrix $W \in \mathbb{R}^{K\times d}$.
Furthermore, we replace the log-likelihood loss with a squared difference loss,
as proposed by~\cite{Mao17}. Hence, we optimise the following revised training objective:
\begin{equation*}\label{eq:gan1}
   \mathcal{L}(\Phi, \D, W) = \Ex_{\by\thicksim \mathcal{Y}}[\D(W^T\by)^2] + \Ex_{\bx \thicksim \mathcal{X}}[(1 - \D(W^T\Phi(\bx)))^2].
\end{equation*}
The embeddings $W$ are trained to aid discrimination among symbols by solving
$\min_{\Phi}\max_{\D,W} \mathcal{L}(\Phi, \D, W)$. Learning such embeddings
improved the speed of convergence and final accuracy, as also noted in~\cite{Gomez18},
while using square differences improved numerical stability.\\

\parag{Discussion: Why is this a feasible learning problem?} While learning to
recognise visual symbols without any paired data seems unattainable, the tight
structure of natural language provides sufficient constraints to enable learning.
First, lexically valid text strings form a tiny sub-space of all possible
permutations of symbols, \eg there are only ${\approx}13k$ valid English words of
length 7, as opposed to almost 8 billion permutations of the 26 English letters.
Second, the relative frequencies of the characters and their co-occurrence patterns
impose further constraints (see~\cref{sec:learn-order} for correlation between
character-frequency and learning).
These constraints combined with strong correlation between the input image and
the predicted characters are sufficient to drive learning successfully.

\begin{figure*}[t]
  \centering
  \includegraphics[width=0.97\textwidth]{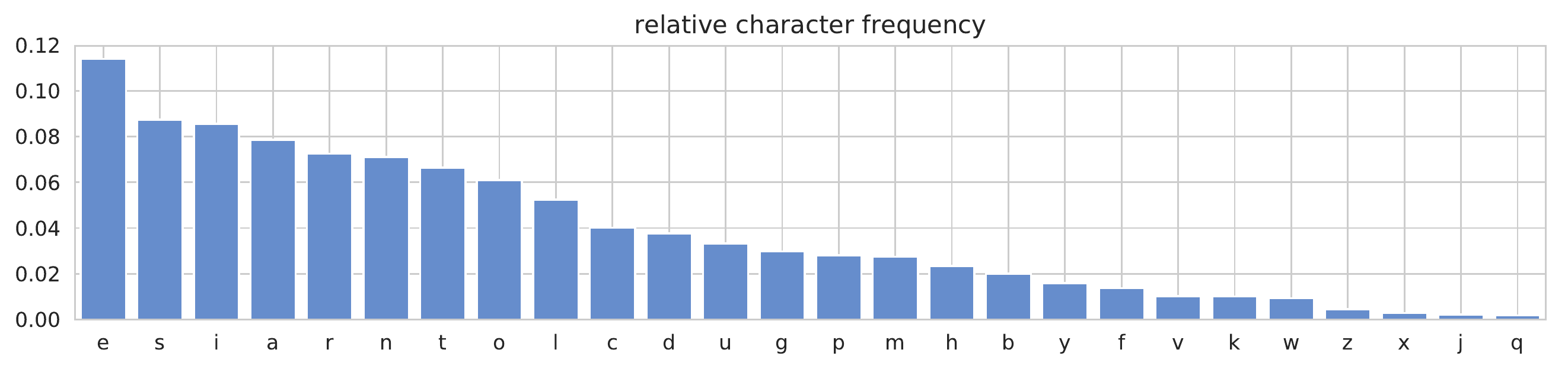}\vspace{-1mm}
  \includegraphics[width=\textwidth]{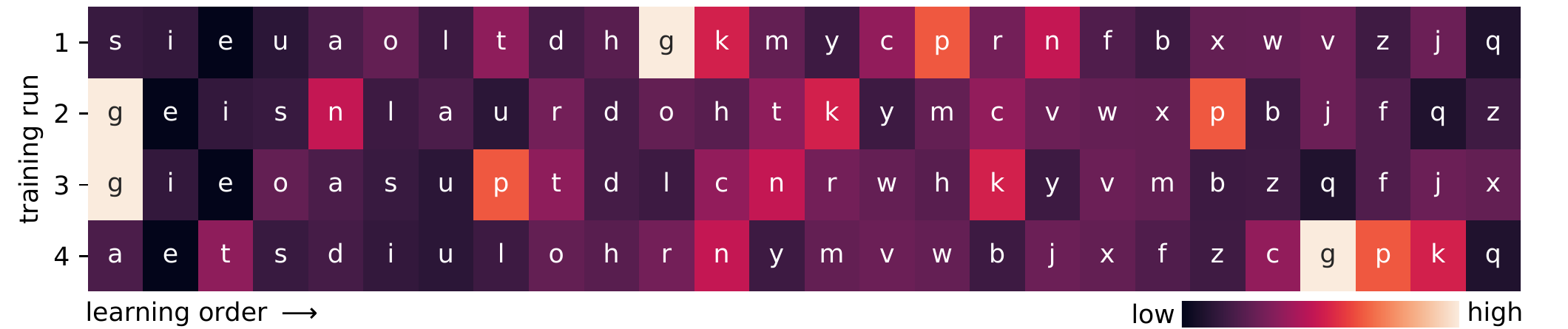}
  \caption{Learning order for different characters.
  \textmd{The order in which the various characters are learnt is strongly correlated
  (Spearman's rank correlation coefficient $\rho = 0.80$, p-value $< 1e{-}5$)
  to their frequency in the English language  \textbf{[top]}.
  Ranking for the learning order is based on the training iteration number
  at which the model achieves $50\%$ accuracy for a given character.
   \textbf{[bottom]} Rankings from four different training runs are presented to show the variance
  --- bright colours signify high variance in rank across runs,
  while dark colours correspond to low variance.
  The character \texttt{\{g\}} is a curious exception to the trend, as it is sometimes
  learnt first (runs $2, 3$);
  see~\cref{sec:learn-order} for the reason and further discussion.}}
  \label{fig:torder-var}
\end{figure*}

\section{Implementation}\label{sec:impl}

Both, the recogniser ($\Phi$) and the discriminator ($\D$) are implemented as fully-convolutional networks~\cite{Long15}.
The recogniser ingests an image of text and produces a sequence of character logits.
The discriminator operates instead on character strings
represented as sequence of character vectors, and produces a scalar discrimination score as output.
The discriminator acts as a \emph{spell-checker}, pointing out the errors
in the generated strings.
We describe their architecture and optimisation details below.\\

\begin{figure}[b]
  \centering
  \includegraphics[width=\linewidth]{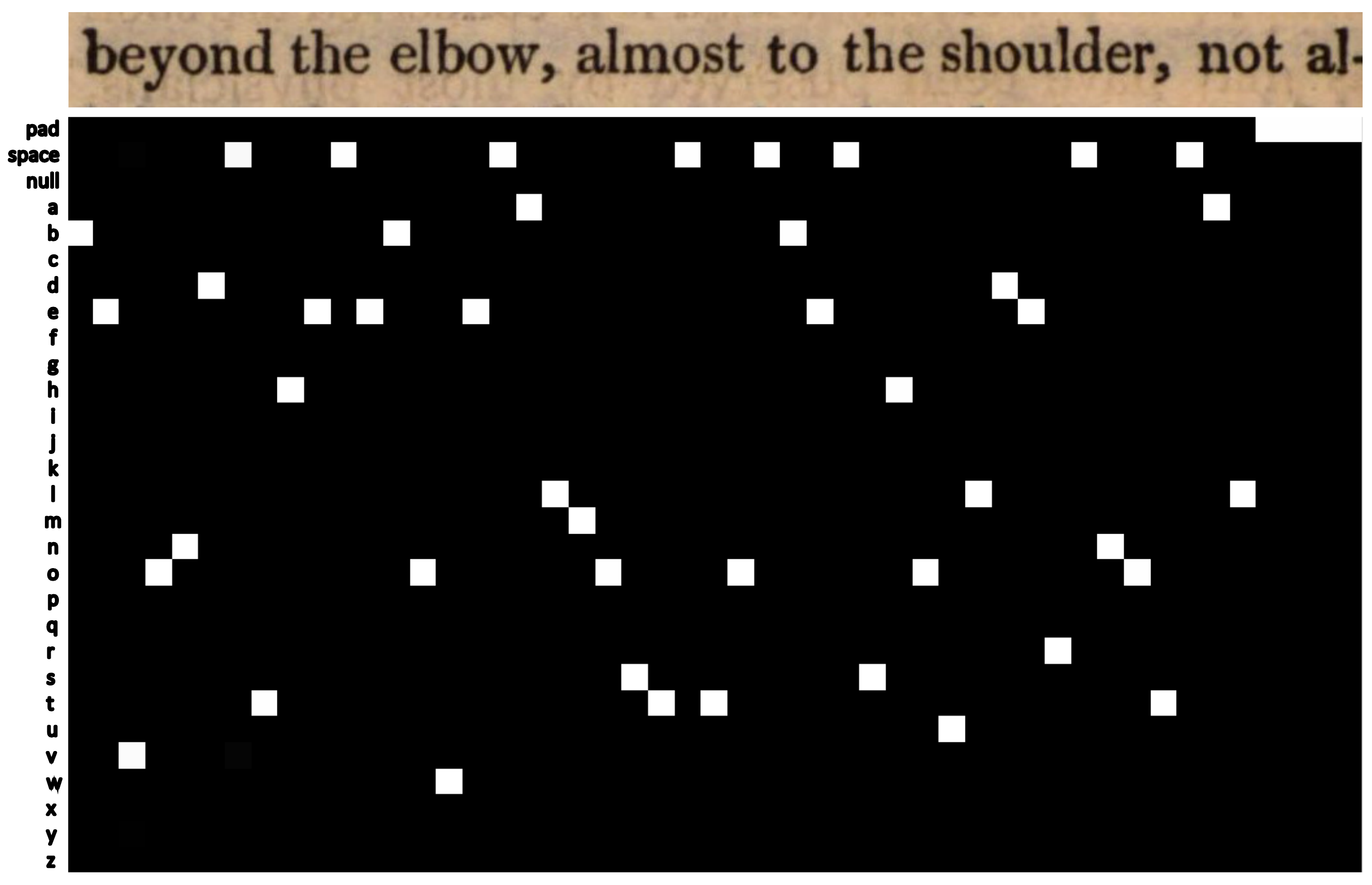}
  \vspace{-5mm}
  \caption{Character sequence representation.
   \textmd{Text strings are represented as sequences of $n$ one-hot (for \emph{real} strings)
   or $\operatorname{SoftMax}$ normalised logits (for \emph{predictions}) over
   $|\mathcal{A}| = K$ character classes.
   A sample image and the model's prediction are visualised above
   (one-hot \emph{real} strings look similar); here $K = 29$ and $n = 50$.}}
  \label{fig:tickertape}
  \vspace{-6mm}
\end{figure}

\parag{Recogniser $\Phi$.}
We train our models for strings of a maximum fixed number of characters $=n$.
To this end, the input image dimensions are held fixed at $32 \times (n\cdot 2^4)$ pixels
($= \text{height} \times \text{width}$).
Hence, an image of size $H\times W$ is scaled to
$H'\times W' = 32\times \min(\lceil W\cdot\frac{32}{H}\rceil, n\cdot 2^4)$;
if the $W' < n\cdot 2^4$, it is padded on the right with the mean channel intensity.
The recogniser employs four blocks, each consisting of two convolution layers,
followed by a $2\times 2$ max-pooling layer.
Each convolutional layer comprises of 32 filters of $3\times 3$ dimensions,
and is followed by batch-normalisation~\cite{Ioffe15} and leaky-ReLU activation
(slope$=0.2$)~\cite{Maas13}.
Since max-pooling in each block downsamples the input by a factor of two,
final output dimensions are $2\times n \times D$ (where, $D=32$ is the number of features).
The height is collapsed using average-pooling, and each of the $n$ $D$-dimensional
feature vectors are mapped to $|\mathcal{A}| = K$ dimensional logits
through linear projection, yielding a $K \times n$ dimensional tensor.
Note the receptive field of the final (prediction) layer is small to encourage \emph{locality};
specifically, it is $76$ pixels wide which corresponds to ${\approx}2.5$ characters in the image.
Although we train our recogniser on fixed-length strings, yet it generalises to
different other lengths due to its fully-convolutional architecture (see~\cref{sec:wlen-generalise}).\\

\begin{figure}[t]
  \centering
  \includegraphics[width=0.97\linewidth]{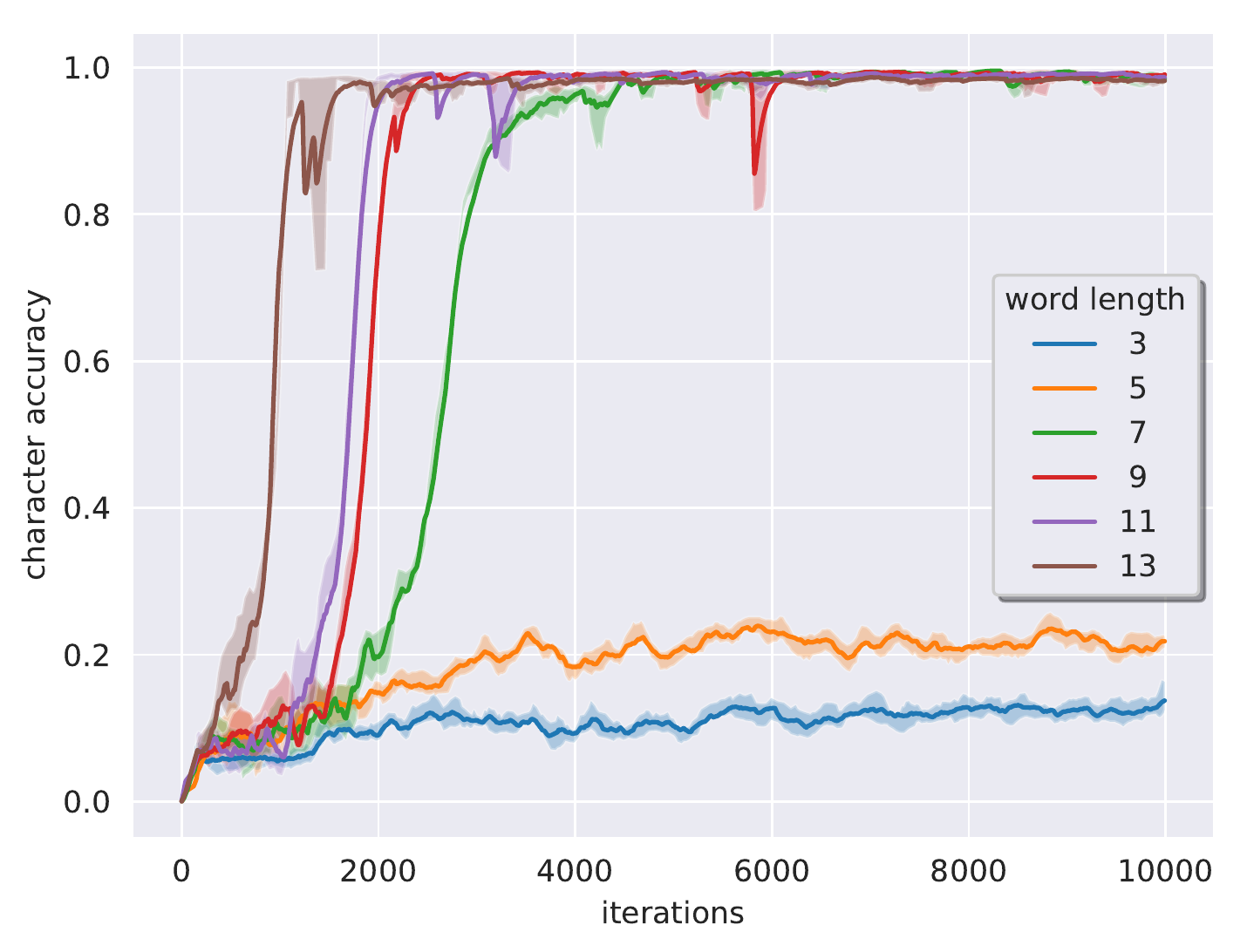}
  \vspace{-5mm}
  \caption{Effect of text length on convergence. \textmd{Training with longer
   words leads to faster convergence: the order of convergence
   \texttt{\{13,11,9,7\}} mirrors the word lengths (see~\cref{sec:wlen}).
   No convergence is seen for models trained on shorter words of length 3 and 5.
   For each word-length, the run with largest area-under-curve (AUC)
   from eight trials is plotted.}}
  \label{fig:wlenconv}
  \vspace{-4.5mm}
\end{figure}

\parag{Discriminator (or spell-checker) $\D$.}
The input $\by$ to the discriminator are $K\times n$ dimensional tensors of
predicted and real strings containing $n$ characters, represented as logits and
one-hot vectors, respectively (see~\cref{fig:tickertape}).
The predicted logits are first normalised through $\operatorname{SoftMax}$ to
a valid probability distribution over the $K$ characters for each of the $n$ positions.
Next, embeddings $\by_e \in \mathbb{R}^{d\times n}$ for both, the real and predicted strings are obtained:
$\by_e  = W^T \by $, where $W\in \mathbb{R}^{K\times d}$ are
the character embeddings ($d = 256$).
We adopt the fully-convolutional \emph{PatchGAN} discriminator architecture~\cite{Isola17,Li16,Liu17ii}, where patches correspond to sub-strings here.
The embedded input $\by_e$ is fed to a stack of five 1D-convolutional layers,
each with~512 filters of size~$5$.
This amounts to a final receptive field of~21 characters which helps to enforce long-range structure.
Each layer is followed by layer-normalisation~\cite{Ba16} and leaky-ReLU ($\text{slope} = 0.2$);
zero padding is used to maintain the size.
The resulting $d \times n$ dimensional output is linearly projected to $1 \times n$,
and average-pooled to obtain the final scalar score $\D(\by)$.\\

\parag{Optimization details.}
Recogniser, discriminator and character embeddings are trained jointly end-to-end.
The parameters are initialised with Xavier initialization~\cite{Glorot10}.
We use the RMSProp optimizer~\cite{Tieleman12} with a constant learning rate of 0.001.
The two-part discriminator loss objective is multiplied with $\frac{1}{2}$ as in~\cite{Isola17}.
The models are implemented in TensorFlow~\cite{Abadi16}.


\section{Experiments}\label{sec:exp}
Our experiments have two primary goals.
First, is an extensive analysis of various factors which affect the training:
we study ---
(1)~the impact of the length of training sequences on convergence (\cref{sec:wlen});
(2)~the order in which various characters are learnt and its correlation with
their frequencies (\cref{sec:learn-order}),
(3)~generalisation ability of the fully-convolutional recogniser to different
sequence lengths (\cref{sec:wlen-generalise}), and
(4) impact of varying the text corpus on recognition accuracy (\cref{sec:corpus}).
For these experiments we use synthetically generated text data
as it provides fine control over various nuisance factors.
The second objective is to show applicability of the proposed method
to \emph{real} document images (\cref{sec:real}).
We first describe the datasets used in our experiments in \cref{sec:data},
and then present the results.

\subsection{Datasets}\label{sec:data}
\parag{Synthetic data.} We generate synthetic text data to simulate
old printed documents. Synthetic data aids the controlled ablation studies,
as it provides tight control over the various factors,
\eg text content, font style and glyph geometry, background, colours, and other
noise parameters.
We sample the text content from two different sources depending on the
experimental setting --- (1) \emph{words}:
individual English words are sourced from a lexicon of 90K words used in the Hunspell
spell-checker~\cite{hunspell}, and
(2) \emph{lines}: these are full valid English language text strings
extracted from the 2011 news-crawl corpus provided by WMT~\cite{wmt}.
Note, these text sources are used for rendering images, as well as
for providing examples of valid strings to the discriminator.
To limit the variance in position of characters, we use the \texttt{VerilySerifMono}
fixed-width font. The background image data is sampled from the margins of historical
books~\cite{Antonacopoulos13} to simulate various noise effects.
The font colour is sampled from a $k$-means colour model learnt from the same dataset.
The character set consists of the 26 English letters, one space character,
and one additional \texttt{null} class for padding smaller strings,
\ie $|\mathcal{A}| = K = 28$. Punctuations, and other symbols in the text are
ignored; lower and upper case letters are mapped to the same class.
Different synthetic datasets are generated as required by the experiments;
the training sets consist of 100k image samples, while the tests set contain 1k samples.
\Cref{fig:synth} visualises some synthetically generated samples.\\

\begin{figure}[t]
  \centering
  \includegraphics[height=38mm]{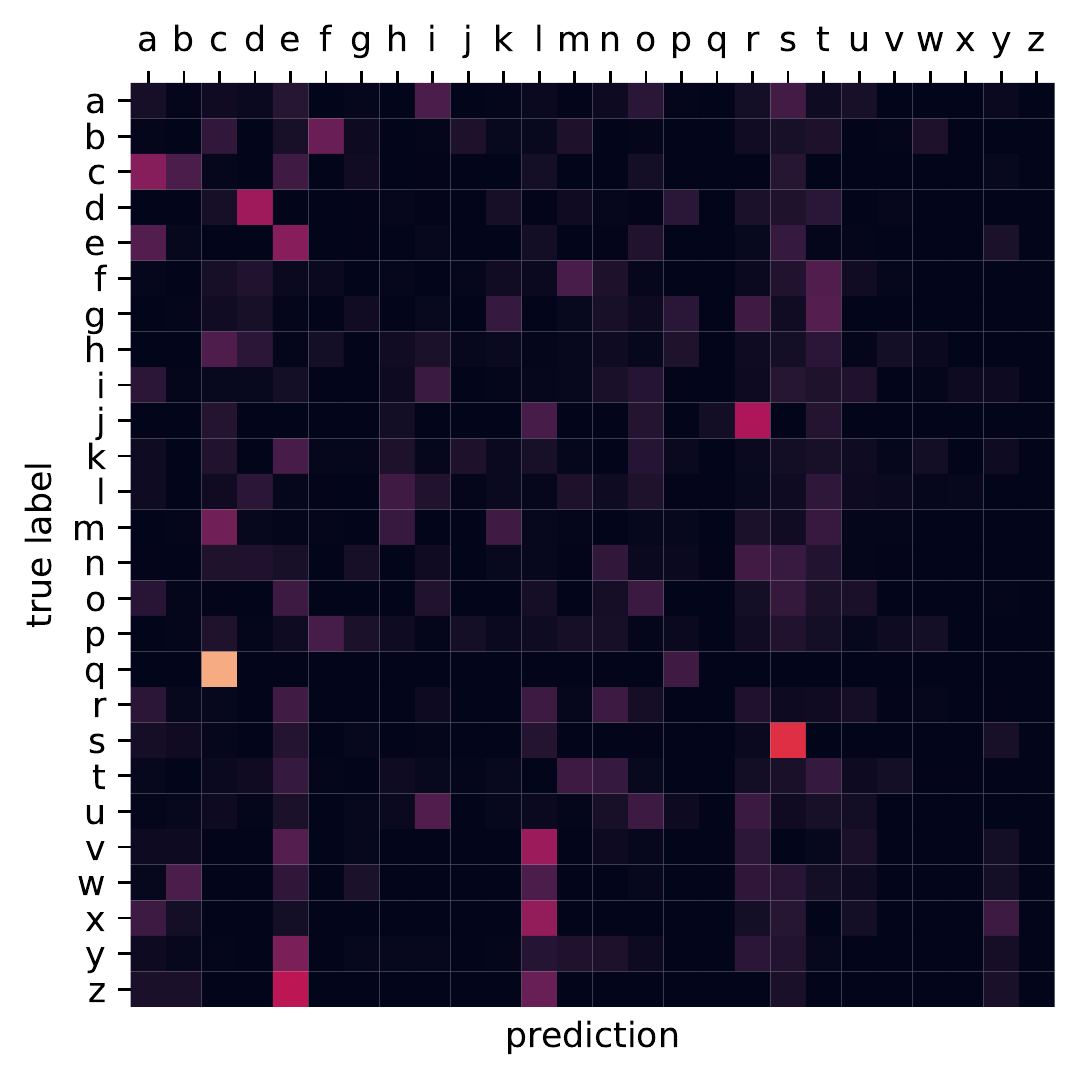}
  \includegraphics[height=38mm]{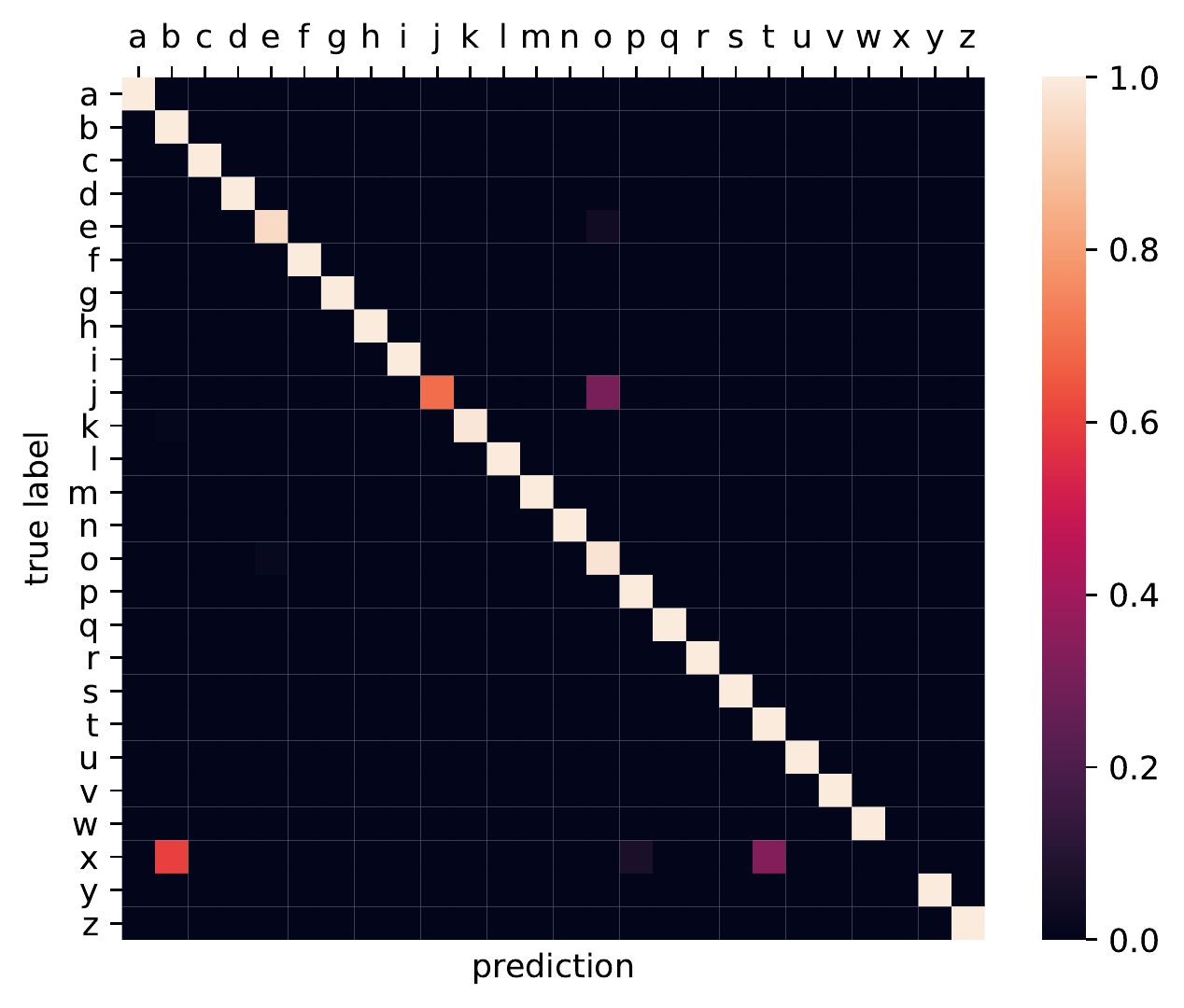}
  \vspace{-3mm}
  \caption{Confusion matrices for models trained on words of length 5 \& 7.
    \textmd{The model trained on length-5 words does not converge to a high accuracy,
    while the one trained on length-7 words does (see~\cref{fig:wlenconv} and
    \cref{sec:wlen}).
    Further, the accuracy for a character depends on its frequency:
    the length-5 model \textbf{[left]} confuses most characters,
    yet it is quite accurate for the common character \texttt{\{s\}};
    while, the length-7 model \textbf{[right]} recognises most characters with
    high accuracy, yet it confuses the two least common characters \texttt{\{j,x\}}
    (see~\cref{sec:learn-order}).}}
  \label{fig:confmat}
\end{figure}

\begin{figure}[t]
  \centering
  \includegraphics[width=\linewidth]{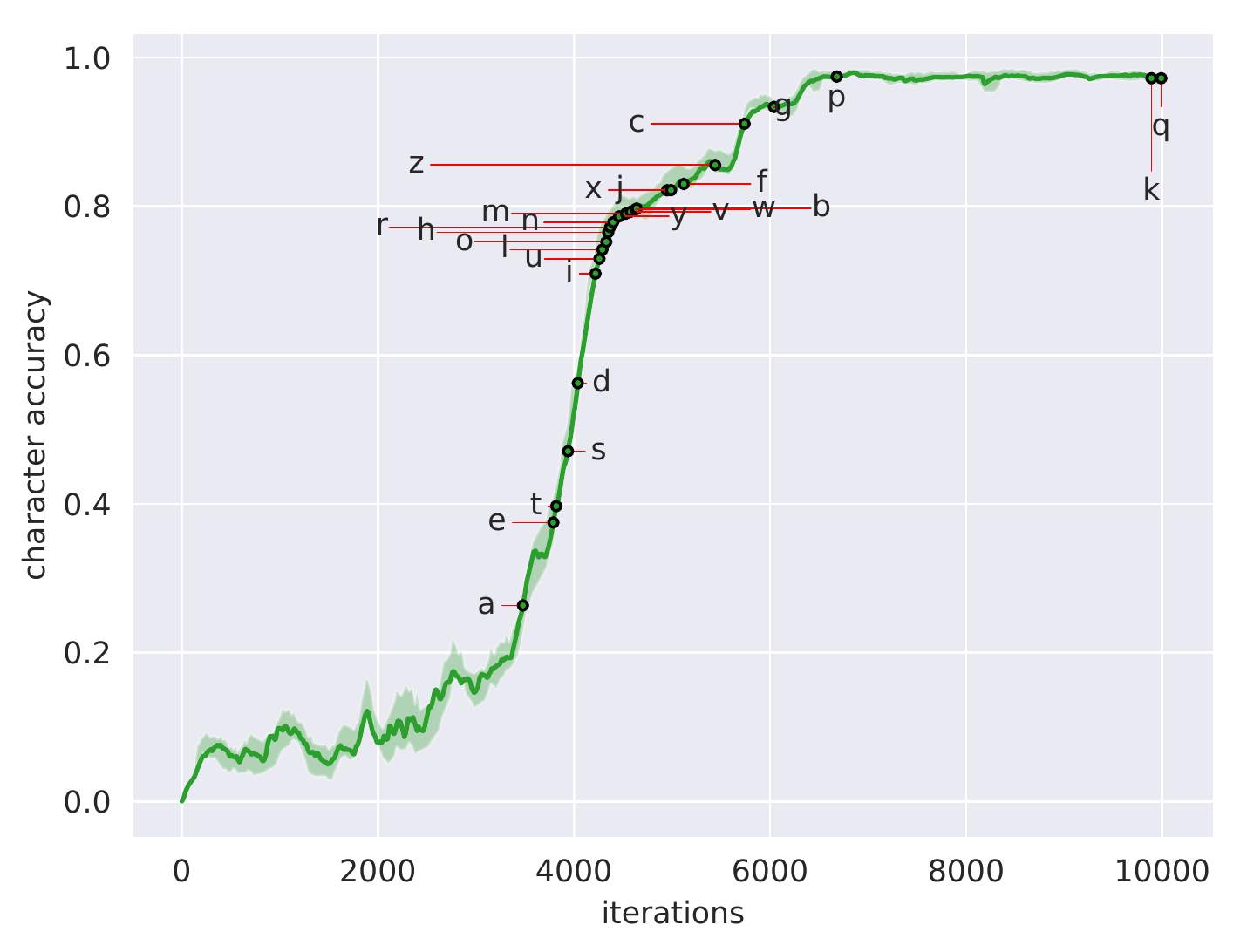}
  \vspace{-5mm}
  \caption{Temporal learning order.
  \textmd{Training iterations for a model trained on length-7 words,
   annotated at the steps when it becomes at least $50\%$ accurate for each character.
   This curve corresponds to \texttt{run-\#4} in~\cref{fig:torder-var}.
   The characters are learnt in the order of their frequencies
   (see~\cref{sec:learn-order}).}}
  \label{fig:torder}
  \vspace{-4mm}
\end{figure}

\parag{Real data.} For testing the validity of our method on real text images,
we use a scanned historical printed book from the
Google1000 dataset~\cite{Google1k}. Specifically, we use the book titled
\emph{Observations on the Nature and Cure of Gout} by James Parkinson~\cite{Parkinson1805}.
For simplicity, we discard cover, title and start-of-chapter pages, and pages
with significant number of footnotes; we only work with the remaining 140 pages
(total 200 pages) which contain text in a relatively uniform font.
Nevertheless, this data is still challenging due to:
(1)~non-fixed-width font which makes character segmentation difficult,
(2)~varying spacing between words due to fully justified alignment,
(3)~varying case (lower/upper) and italics,
(4)~different background colours and textures,
(5)~show-through from the back of the page,
(6)~fading and other noise elements, and
(7)~presence of various punctuations and other symbols.
We use the localisation output of the provided OCR engine output to
segment the pages into lines;
first 300 lines are assigned to the test set, while the remaining
3000 form the training set (no page is shared between the splits).
We use the provided OCR text output for lines in the training split,
as examples of valid text strings for the discriminator.
Note, these strings are sampled uniformly at random during training,
and hence, do not have any direct correspondence to images in the training batch.
The text lines typically consist of ${\approx}50$ characters.
The character-set consists of 26 English letters, one space character,
one unknown \texttt{<UNK>} character, and one \texttt{null} class for padding,
for a total of $|\mathcal{A}| = K = 29$ characters.
We do not distinguish between upper and lower cases; the following
symbols and punctuations: \texttt{, . ? ! ` " * ( )} are suppressed (ignored),
and any other character is mapped to \texttt{<UNK>}.
\Cref{fig:real} visualises some sample text-lines.\\\\

\parag{Metrics.} We measure accuracy at the \emph{character} and \emph{word} levels:
\begin{itemize}
  \item\textbf{character accuracy}: this is computed as\\
  $1 - \dfrac{1}{N}\sum\limits_{i=1}^N \frac{\operatorname{EditDist}\left(\by_{\text{gt}}^{(i)}, \by_{\text{pred}}^{(i)}\right)}{\operatorname{Length}\left(\by_{\text{gt}}^{(i)}\right)}$,
  where $\by_{\text{gt}}^{(i)}$ and $\by_{\text{pred}}^{(i)}$ are the $i^{\text{th}}$ ground-truth and predicted
  strings respectively in a dataset containing $N$ strings;
  $\operatorname{EditDist}$ is the \emph{character-level} \emph{Levenshtein} distance~\cite{Levenshtein66};
  $\operatorname{Length}\left(\by_{\text{gt}}\right)$ is the number of characters in $\by_{\text{gt}}$.
  \item \textbf{word accuracy}: computed as \emph{character accuracy} above, but
  here the \emph{Levenshtein} distance uses \emph{words} (contiguous strings
  demarcated by space) as tokens, and is normalized by number of ground-truth words in $\by_{\text{gt}}$.
\end{itemize}

\begin{figure}[t]
  \centering
  \includegraphics[width=\linewidth]{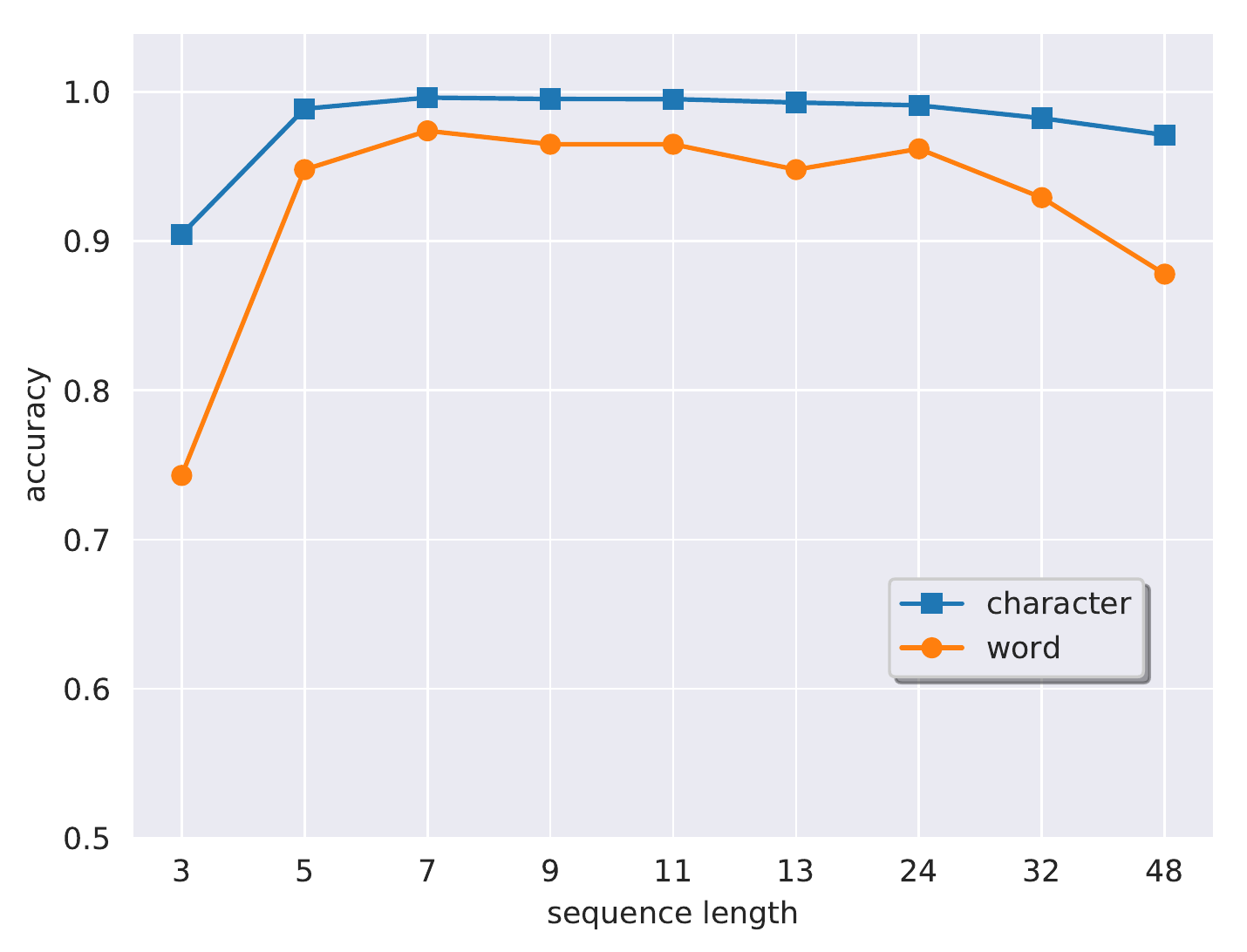}
  \vspace{-5mm}
  \caption{Generalisation to different sequence lengths.
  \textmd{A model trained on text-strings of length 24,
  is evaluated on images containing both shorter and longer strings of lengths ---
  $\{3,5,7,9,11,13,32,48\}$.
  Word and character accuracies are plotted.
  The fully-convolutional architecture of the recognition network
  enables significant generalisation to lengths not in the training set,
  with small variance in performance (see~\cref{sec:wlen-generalise}).}}
  \label{fig:wlen-eval}
\end{figure}

\subsection{Effect of text length on convergence}\label{sec:wlen}
Although earlier works use low-order, namely uni/bi-gram statistics for alignment
~\cite{Lee02,Kozielski14}, higher-order $n$-grams could be more informative.
In this experiment we examine the impact of the \emph{length} of the training
text-sequences on convergence.
We train separate models on synthetic datasets containing \emph{one} word of a given length,
namely --- \texttt{\{3,5,7,9,11,13\}}.
\Cref{fig:wlenconv} tracks \emph{character accuracy} as the training progresses;
due to instabilities in training GANs, we train on each word-length eight times, and
plot the run with the maximum area-under-curve (AUC) (earliest ``take-off'').
Note, models trained on longer words converge faster, achieving ${\approx} 99\%$ \emph{character~accuracy}.
In detail, the model trained on length-13 words converges the fastest,
followed by those trained on 11, 9, and 7 (in order).
No convergence is seen for shorter lengths 3 and 5 (although the accuracy is higher for 5).
This confirms that longer text-sequences impose stronger structural constraints on
the possible outputs, leading to faster convergence.
\Cref{fig:confmat} visualises the confusion matrices for models trained
on lengths 5 and 7; the length-5 model confuses most characters,
whereas the length-7 model recognises most characters almost perfectly.
Note, convergence is independent of the \emph{number of distinct word-instances}
for a given length:
there are more length-5 words (${\approx} 7000$) than length-13 (${\approx} 2500$)
in the lexicon, yet training with length-5 words does not converge.
Further, words of length 7 are the most in number (${\approx}13000$),
yet it converges last.

\begin{figure*}
  \centering
  \resizebox{0.95\textwidth}{!}{
  \includegraphics[height=58mm]{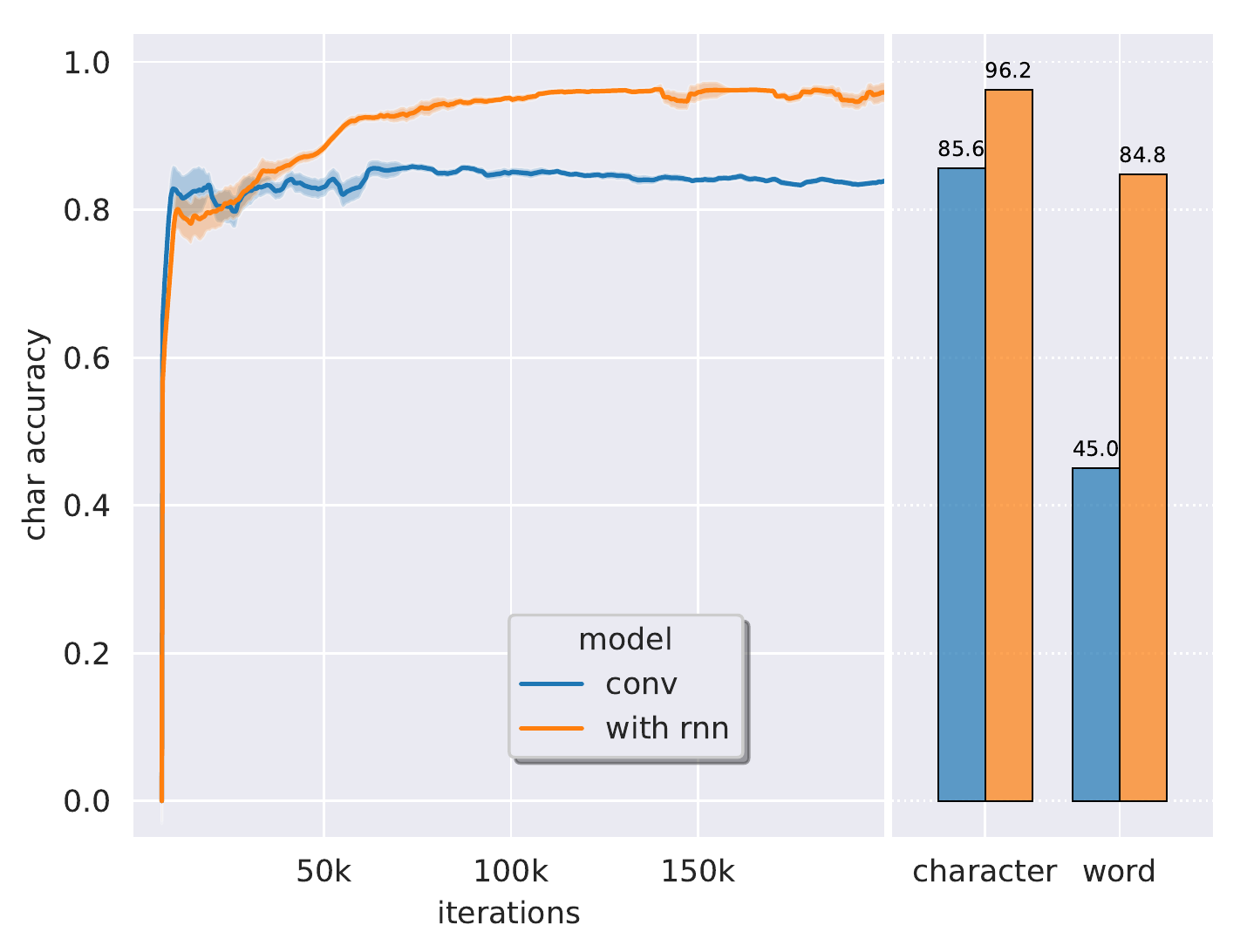}
  \includegraphics[height=58mm]{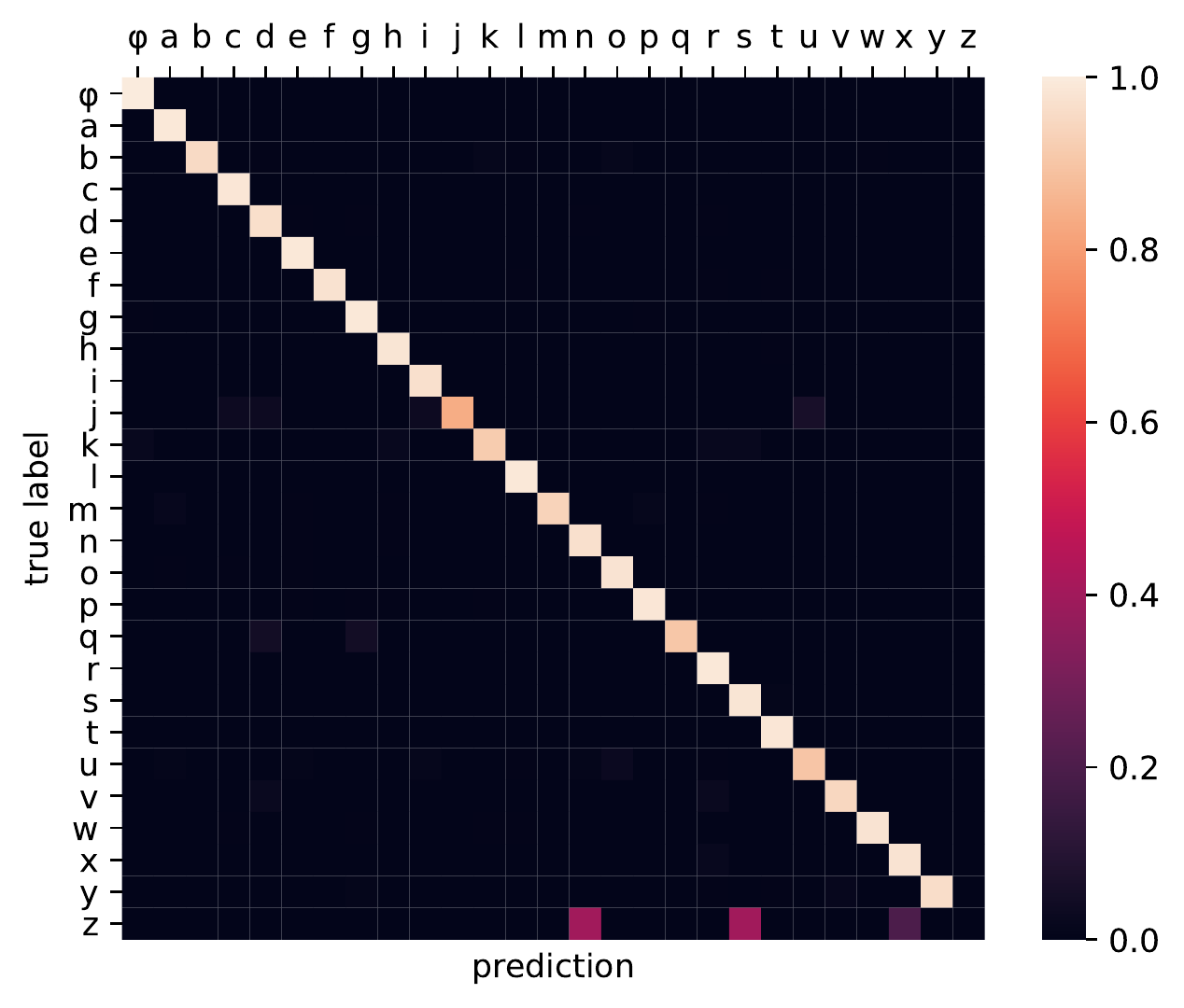}}
  \vspace{-3mm}
  \caption{Recognising historical printed books.
    \textmd{\textbf{[left]} \emph{Character \& word accuracy}
    on the test split of the \emph{real} dataset (see~\cref{fig:real})
    for both the fully-convolutional and the \emph{skip-RNN} recognition models.
    \emph{Skip-RNN} dramatically improves: \emph{character~accuracy} from $85.6\%$ to $96.2\%$,
    and \emph{word~accuracy} from $45.0\%$ to $84.8\%$.
    \emph{Character accuracy} on the test-set is also visualised against the training iterations.
    \textbf{[right]}~Confusion matrix on the test set:
    all characters are predicted with high accuracy, except for the low-frequency~\texttt{\{z\}}.
    \{$\upvarphi$\} stands for the \texttt{\{space\}} character (see~\cref{sec:real}).}}
  \label{fig:real-acc}
\end{figure*}

\subsection{Which character is learnt first?}\label{sec:learn-order}
We examine the dynamics of learning, more specifically, we probe the order
in which the model learns about different symbols --- is there a pattern?
\Cref{fig:torder-var} visualises the order in which models (trained on
synthetic word images of length 7) achieve an accuracy of at least $50\%$ for each character.
We note that this ranking is highly correlated with the frequency of the characters
in the English language --- Spearman's rank correlation coefficient $\rho = 0.80$,~p-value~$< 1e{-}5$.
It further visualises the variance in the ranking of the characters across
multiple runs. The characters at the extremities of the frequency distribution
have low variance --- common characters (\eg \texttt{e,s,i,a}) are almost always
learnt first, and the least common characters (\eg \texttt{z,x,j,q}) are learnt
last; while characters in the middle, \viz \texttt{\{g,p\}} show the highest variance.
The character \texttt{\{g\}} is a curious exception as it is sometimes learnt first.
This is because of $8.54\%$ of the training (length-7) words end in the suffix \texttt{`-ing'}.
Hence, \texttt{\{g\}} appears at the last position quite frequently, and becomes
relatively easy to learn.
\Cref{fig:torder} annotates the training steps at which the accuracy for a character
first reaches $50\%$. After the model becomes confident about the first symbol \texttt{\{a\}},
it quickly learns the other most commons ones; then it slowly learns
the less frequent symbols in the order of their frequencies.
Further, \cref{fig:confmat} visualises the confusion-matrices for models
trained on word-lengths 5 and 7. Again, we can note the dependence on
character frequencies --- even though model for length-5 words does not converge
(see~\cref{sec:wlen}), it is somewhat accurate about the frequent character \texttt{\{s\}},
while the length-7 model is almost perfect at recognising most characters,
yet it confuses two of the least common characters \texttt{\{j,x\}}.

\subsection{Generalisation to different lengths}\label{sec:wlen-generalise}
The fully-convolutional architecture of our recognition network generalises
to images of lengths significantly different from those it was trained on.
To demonstrate this, we train a model on synthetic \emph{text-strings} of
length 24 (containing multiple words), and evaluate on synthetic images of different lengths:
(1) \emph{shorter} single-word images of lengths --- $\{3,5,7,9,11,13\}$, and
(2) \emph{longer} text-string (multiple words) images of lengths --- $\{32, 48\}$.
\Cref{fig:wlen-eval} plots the recognition accuracy against the word lengths.
We note excellent and consistent \emph{character} (${\approx}99\%$)
and \emph{word} accuracies (${\approx} 95\%$) for both, shorter and longer lengths (5 --- 32).
Note, this demonstrates significant generalisation ability, as the model
is never trained on such images.
There is a drop in the character accuracy (${\approx}95\%$) for length-48
text-strings, as the model does not learn a long-range language model.
Performance suffers for words of length 3 due to
image-edges being close in short images, which is not encountered during training with
images of long words.

\begin{table}[b]
  \begin{tabular}{@{}lccc@{}}
    \toprule
    \multirow{2}{*}{corpus $\rightarrow$} & \multirow{2}{*}{(1) WMT} & (2) WMT        & \multirow{2}{*}{(3) War \& Peace} \\
                                           &                      & no overlap &                               \\ \midrule
    char                                   & 98.98                & 99.13      & 98.43                         \\
    word                                   & 95.33                & 96.08      & 92.53                         \\ \bottomrule
  \end{tabular}
  \vspace{3mm}
  \caption{Effect of varying the text corpus on recognition accuracy.
  \textmd{Text-strings are sampled from three increasingly distant text corpora.
  This has a small adverse effect on recognition \emph{word} and \emph{character} accuracies
  (in $\%$) (see~\cref{sec:corpus}).}}
  \label{tab:corpus}
\end{table}

\subsection{Varying the text corpus}\label{sec:corpus}
We examine the impact of varying the text corpus from which samples of text strings
are obtained, on the recognition accuracy.
We examine the following three different sources for sampling the strings.
The synthetic text-images are held constant across the three settings, and
contain up to 24 characters with the text content in them sampled from
WMT newscrawl (as before).
(1)~same strings as in text-images but randomly sampled for each batch,
(2)~strings from the same corpus (WMT newscrawl) but with no overlap with
text-image strings, and
(3)~strings sampled from a very different corpus, namely, Tolstoy's \emph{War and Peace}.
\Cref{tab:corpus} summarizes the \emph{character} and \emph{word} accuracies.
Training with the completely unrelated lexicon ($\#3$) does have a small adverse effect
(\emph{word} accuracy drops to $92.53\%$ from $\approx 95\%$), while using a related
lexicon $(\#2)$ does not have such an effect.

\begin{figure}[t]
  \centering
  \includegraphics[width=0.98\linewidth]{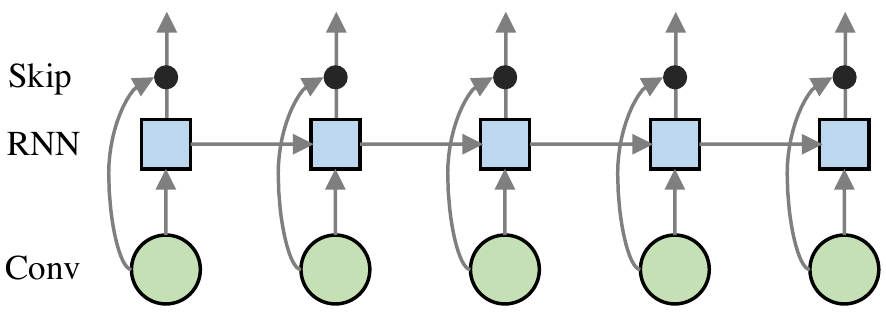}
  \vspace{-3mm}
  \caption{Skip-RNN architecture for real text images.
   \textmd{Non-uniform spacing and non-fixed width fonts pose a significant
   challenge to the fully-convolutional recogniser.
   We augment the recognition network with a \emph{skip-RNN}, which acts on the
   convolutional features, and predicts residual updates to the inputs
   (the residual predictions are \emph{added} to the inputs).
   This improves the \emph{word accuracy} from ${\approx}45\%$ to ${\approx}85\%$
   (see~\cref{fig:real-acc}).}}
  \label{fig:skip-rnn}
\end{figure}

\subsection{Recognising a historical printed book}\label{sec:real}
Finally, we apply our model to \emph{real} text-line images extracted
from a historical printed book (see \cref{sec:data} for dataset details).
As noted in \cref{sec:data}, non-fixed width fonts and fully-justified text alignment introduce
non-uniform spacing between characters and words. This poses a significant
challenge to the fully-convolutional recogniser, making segmentation of the
text-image into individual characters difficult (see \cref{fig:real} for example images).
Hence, we augment the penultimate layer of the fully-convolutional recogniser
with a \emph{skip-RNN} --- a \emph{uni}-directional (left to right)
RNN (256-dimensional LSTM) with a residual skip-connection~\cite{He15} (see~\cref{fig:skip-rnn}).
The RNN lends pliability to the convolutional features, thereby aids character segmentation.
All other model parameters are as those used for the synthetic data experiments
(see \cref{sec:impl}), except: (1)~the discriminator
filter size is increased from 5 to 11, and (2)~number of layers is doubled to 8
to exploit the long-term structure in the much longer text strings
(${\approx}50$ characters each).
\Cref{fig:real-acc} visualises the \emph{character} and \emph{word} accuracies for
both the fully-convolutional and \emph{skip-RNN} recognition models.
\emph{Skip-RNN} dramatically improves the recognition performance:
\emph{word~accuracy} improves from $45.0\%$ to $84.8\%$, while
\emph{character~accuracy} improves from $85.6\%$ to $96.2\%$.
\Cref{fig:real} visualises randomly selected examples from the test set
and shows the model's predictions;
the predictions are comparable to the ``ground-truth'' annotations
obtained from Google's OCR engine.
\Cref{fig:real-acc} also visualises the confusion matrix for the character classes:
all characters are predicted with high accuracy, except for the low-frequency~\texttt{\{z\}}.
Full page read-outs from our model are visualised in \cref{sec:appendix}.

\section{Conclusion}\label{sec:conclude}
We have developed a method for training a text recognition network
using only \emph{unaligned} examples of text-images and valid text strings.
We have presented detailed analysis for various aspects of the proposed method.
We have established ---
(1) positive correlation between the length of the input text and convergence rates;
(2) the order in which the characters are learnt is strongly dependent on their
relative frequencies in the text;
(3) the generalisation ability of our method to input images of different lengths,
specifically our recognition model trained on strings of length 24 generalises
to both much shorter and longer strings (3 -- 48) without drastic degradation in performance;
(4) the effect of varying the text corpus used as the source of valid sentences
on the recognition accuracy.
Finally, we have shown successful recognition on real text images, without using
any labelled supervisory data.
These results open up a new and promising direction for training sequence recognition models
for structured domains (\eg language) given no labelled training data.
The proposed method is applicable not just to text images,
but other modalities as well, \eg speech and gestures.

\begin{acks}
  We thank Iasonas Kokkinos, Triantafyllos Afouras, and Weidi Xie for insightful discussions,
  and anonymous reviewers for their detailed feedback.
  Financial support was provided by \grantsponsor{AIMS}{UK EPSRC AIMS CDT}{} \grantnum{AIMS}{EP/L015987/2},
  \grantsponsor{SEEBIBYTE}{EPSRC Seebibyte Grant}{} \grantnum{SEEBIBYTE}{EP/M013774/1}, and \grantsponsor{CLARENDON}{Clarendon Fund scholarship}{}.
\end{acks}

\bibliographystyle{abbrvnat}
\bibliography{bib/refs,bib/vgg_local,bib/vgg_other,bib/shortstrings,bib/bmvc18}

\begin{thebibliography}{81}
\providecommand{\natexlab}[1]{#1}
\providecommand{\url}[1]{\texttt{#1}}
\expandafter\ifx\csname urlstyle\endcsname\relax
  \providecommand{\doi}[1]{doi: #1}\else
  \providecommand{\doi}{doi: \begingroup \urlstyle{rm}\Url}\fi

\bibitem[wmt(2018)]{wmt}
{EMNLP} conference on machine translation, 2018.

\bibitem[Abadi et~al.(2016)Abadi, Barham, Chen, Chen, Davis, Dean, Devin,
  Ghemawat, Irving, Isard, et~al.]{Abadi16}
M.~Abadi, P.~Barham, J.~Chen, Z.~Chen, A.~Davis, J.~Dean, M.~Devin,
  S.~Ghemawat, G.~Irving, M.~Isard, et~al.
\newblock Tensorflow: a system for large-scale machine learning.
\newblock In \emph{OSDI}, volume~16, pages 265--283, 2016.

\bibitem[Aldarrab(2017)]{Aldarrab17}
N.~Aldarrab.
\newblock Decipherment of historical manuscripts.
\newblock Master's thesis, University of Southern California, 2017.

\bibitem[Almaz{\'a}n et~al.(2014)Almaz{\'a}n, Gordo, Forn{\'e}s, and
  Valveny]{Almazan14}
J.~Almaz{\'a}n, A.~Gordo, A.~Forn{\'e}s, and E.~Valveny.
\newblock Word spotting and recognition with embedded attributes.
\newblock \emph{IEEE PAMI}, 36:\penalty0 2552--2566, 2014.

\bibitem[Alsharif and Pineau(2014)]{Alsharif14}
O.~Alsharif and J.~Pineau.
\newblock End-to-end text recognition with hybrid {HMM} maxout models.
\newblock In \emph{Proc. {ICLR}}, 2014.

\bibitem[Antonacopoulos et~al.(2013)Antonacopoulos, Clausner, Papadopoulos, and
  Pletschacher]{Antonacopoulos13}
A.~Antonacopoulos, C.~Clausner, C.~Papadopoulos, and S.~Pletschacher.
\newblock {ICDAR} 2013 competition on historical book recognition (hbr 2013).
\newblock pages 1459--1463. IEEE, 2013.

\bibitem[Artetxe et~al.(2017)Artetxe, Labaka, Agirre, and Cho]{Artetxe17}
M.~Artetxe, G.~Labaka, E.~Agirre, and K.~Cho.
\newblock Unsupervised neural machine translation.
\newblock In \emph{Proc. {ICLR}}, 2017.

\bibitem[Ba et~al.(2016)Ba, Kiros, and Hinton]{Ba16}
J.~L. Ba, J.~R. Kiros, and G.~E. Hinton.
\newblock Layer normalization.
\newblock \emph{arXiv preprint arXiv:1607.06450}, 2016.

\bibitem[Bahdanau et~al.(2015)Bahdanau, Cho, and Bengio]{Bahdanau15}
D.~Bahdanau, K.~Cho, and Y.~Bengio.
\newblock Neural machine translation by jointly learning to align and
  translate.
\newblock In \emph{Proc. {ICLR}}, 2015.

\bibitem[Berg-Kirkpatrick et~al.(2013)Berg-Kirkpatrick, Durrett, and
  Klein]{berg2013unsupervised}
T.~Berg-Kirkpatrick, G.~Durrett, and D.~Klein.
\newblock Unsupervised transcription of historical documents.
\newblock In \emph{Proceedings of the 51st Annual Meeting of the Association
  for Computational Linguistics (Volume 1: Long Papers)}, volume~1, pages
  207--217, 2013.

\bibitem[Bissacco et~al.(2013)Bissacco, Cummins, Netzer, and Neven]{Bissacco13}
A.~Bissacco, M.~Cummins, Y.~Netzer, and H.~Neven.
\newblock {PhotoOCR}: Reading text in uncontrolled conditions.
\newblock In \emph{Proc. {ICCV}}, 2013.

\bibitem[Bunke et~al.(2004)Bunke, Bengio, and Vinciarelli]{bunke04}
H.~Bunke, S.~Bengio, and A.~Vinciarelli.
\newblock Offline recognition of unconstrained handwritten texts using {HMM}s
  and statistical language models.
\newblock \emph{{PAMI}}, 26\penalty0 (6):\penalty0 709--720, 2004.

\bibitem[Casey(1986)]{Casey86}
R.~G. Casey.
\newblock Text {OCR} by solving a cryptogram.
\newblock 1986.

\bibitem[Cho et~al.(2014)Cho, van Merrienboer, Gulcehre, Bahdanau, Bougares,
  Schwenk, and Bengio]{Cho14}
K.~Cho, B.~van Merrienboer, C.~Gulcehre, D.~Bahdanau, F.~Bougares, H.~Schwenk,
  and Y.~Bengio.
\newblock Learning phrase representations using rnn encoder--decoder for
  statistical machine translation.
\newblock In \emph{Proceedings of the 2014 Conference on Empirical Methods in
  Natural Language Processing (EMNLP)}, 2014.

\bibitem[Dempster et~al.(1977)Dempster, Laird, and Rubin]{Dempster77}
A.~P. Dempster, N.~M. Laird, and D.~B. Rubin.
\newblock Maximum likelihood from incomplete data via the {EM} algorithm.
\newblock 39 B:\penalty0 1--38, 1977.

\bibitem[Dooley(2013)]{dooley2013brief}
J.~F. Dooley.
\newblock \emph{A brief history of cryptology and cryptographic algorithms}.
\newblock Springer, 2013.

\bibitem[Glorot and Bengio(2010)]{Glorot10}
X.~Glorot and Y.~Bengio.
\newblock Understanding the difficulty of training deep feedforward neural
  networks.
\newblock In \emph{Proceedings of the thirteenth international conference on
  artificial intelligence and statistics}, pages 249--256, 2010.

\bibitem[Goel et~al.(2013)Goel, Mishra, Alahari, and Jawahar]{Goel13}
V.~Goel, A.~Mishra, K.~Alahari, and C.~V. Jawahar.
\newblock Whole is greater than sum of parts: Recognizing scene text words.
\newblock In \emph{International Conf. on Document Analysis and Recognition
  (ICDAR)}, pages 398--402, 2013.

\bibitem[Gomez et~al.(2018)Gomez, Huang, Zhang, Li, Osama, and Kaiser]{Gomez18}
A.~N. Gomez, S.~Huang, I.~Zhang, B.~M. Li, M.~Osama, and L.~Kaiser.
\newblock Unsupervised cipher cracking using discrete {GAN}s.
\newblock In \emph{Proc. {ICLR}}, 2018.

\bibitem[Goodfellow et~al.(2014)Goodfellow, Pouget-Abadie, Mirza, Xu,
  Warde-Farley, Ozair, Courville, and Bengio]{goodfellow2014generative}
I.~Goodfellow, J.~Pouget-Abadie, M.~Mirza, B.~Xu, D.~Warde-Farley, S.~Ozair,
  A.~Courville, and Y.~Bengio.
\newblock Generative adversarial nets.
\newblock In \emph{Proc. {NIPS}}, 2014.

\bibitem[{Google Inc.}(2018)]{Google1k}
{Google Inc.}
\newblock Book search dataset, Aug 2018.
\newblock Version V.

\bibitem[Gordo(2015)]{Gordo15}
A.~Gordo.
\newblock Supervised mid-level features for word image representation.
\newblock In \emph{Proc. CVPR}, 2015.

\bibitem[Graves et~al.(2006)Graves, Fern{\'a}ndez, Gomez, and
  Schmidhuber]{Graves06}
A.~Graves, S.~Fern{\'a}ndez, F.~Gomez, and J.~Schmidhuber.
\newblock Connectionist temporal classification: labelling unsegmented sequence
  data with recurrent neural networks.
\newblock In \emph{Proceedings of the 23rd international conference on Machine
  learning}, pages 369--376. ACM, 2006.

\bibitem[Gupta et~al.(2016)Gupta, Vedaldi, and Zisserman]{Gupta16}
A.~Gupta, A.~Vedaldi, and A.~Zisserman.
\newblock Synthetic data for text localisation in natural images.
\newblock In \emph{Proc. {CVPR}}, 2016.

\bibitem[He et~al.(2015)He, Zhang, Ren, and Sun]{He15}
K.~He, X.~Zhang, S.~Ren, and J.~Sun.
\newblock Deep residual learning for image recognition.
\newblock \emph{arXiv preprint arXiv:1512.03385}, 2015.

\bibitem[He et~al.(2016)He, Huang, Qiao, Loy, and Tang]{He16DTRN}
P.~He, W.~Huang, Y.~Qiao, C.~Loy, and X.~Tang.
\newblock Reading scene text in deep convolutional sequences, 2016.
\newblock In \emph{The 30th AAAI Conference on Artificial Intelligence
  (AAAI-16)}, volume~1, 2016.

\bibitem[Ho and Nagy(2000)]{Ho00}
T.~K. Ho and G.~Nagy.
\newblock {OCR} with no shape training.
\newblock In \emph{Proc. {ICPR}}, 2000.

\bibitem[Huang et~al.(2007)Huang, Learned-Miller, and McCallum]{Huang07c}
G.~Huang, E.~Learned-Miller, and A.~McCallum.
\newblock Cryptogram decoding for optical character recognition.
\newblock 2007.

\bibitem[Hunspell()]{hunspell}
Hunspell.
\newblock https://hunspell.github.io.

\bibitem[Ioffe and Szegedy(2015)]{Ioffe15}
S.~Ioffe and C.~Szegedy.
\newblock Batch normalization: Accelerating deep network training by reducing
  internal covariate shift.
\newblock In \emph{Proc. {ICML}}, 2015.

\bibitem[Isola et~al.(2017)Isola, Zhu, Zhou, and Efros]{Isola17}
P.~Isola, J.-Y. Zhu, T.~Zhou, and A.~A. Efros.
\newblock Image-to-image translation with conditional adversarial networks.
\newblock In \emph{Proc. {CVPR}}, 2017.

\bibitem[Jaderberg et~al.(2014{\natexlab{a}})Jaderberg, Simonyan, Vedaldi, and
  Zisserman]{Jaderberg14c}
M.~Jaderberg, K.~Simonyan, A.~Vedaldi, and A.~Zisserman.
\newblock Synthetic data and artificial neural networks for natural scene text
  recognition.
\newblock In \emph{Workshop on Deep Learning, NIPS}, 2014{\natexlab{a}}.

\bibitem[Jaderberg et~al.(2014{\natexlab{b}})Jaderberg, Vedaldi, and
  Zisserman]{Jaderberg14}
M.~Jaderberg, A.~Vedaldi, and A.~Zisserman.
\newblock Deep features for text spotting.
\newblock In \emph{Proc. {ECCV}}, 2014{\natexlab{b}}.

\bibitem[Jaderberg et~al.(2015)Jaderberg, Simonyan, Vedaldi, and
  Zisserman]{Jaderberg15a}
M.~Jaderberg, K.~Simonyan, A.~Vedaldi, and A.~Zisserman.
\newblock Deep structured output learning for unconstrained text recognition.
\newblock In \emph{International Conference on Learning Representations}, 2015.

\bibitem[Jaderberg et~al.(2016)Jaderberg, Simonyan, Vedaldi, and
  Zisserman]{Jaderberg16}
M.~Jaderberg, K.~Simonyan, A.~Vedaldi, and A.~Zisserman.
\newblock Reading text in the wild with convolutional neural networks.
\newblock \emph{{IJCV}}, 116\penalty0 (1):\penalty0 1--20, Jan. 2016.

\bibitem[Kae and Learned-Miller(2009)]{Kae09}
A.~Kae and E.~Learned-Miller.
\newblock Learning on the fly: font-free approaches to difficult {OCR}
  problems.
\newblock 2009.

\bibitem[Karatzas et~al.(2013)Karatzas, Shafait, Uchida, Iwamura, Mestre, Mas,
  Mota, Almazan, de~las Heras, et~al.]{Karatzas13}
D.~Karatzas, F.~Shafait, S.~Uchida, M.~Iwamura, S.~R. Mestre, J.~Mas, D.~F.
  Mota, J.~A. Almazan, L.~P. de~las Heras, et~al.
\newblock {ICDAR} 2013 robust reading competition.
\newblock In \emph{Proc. ICDAR}, pages 1484--1493, 2013.

\bibitem[Knight et~al.(2006)Knight, Nair, Rathod, and Yamada]{Knight06}
K.~Knight, A.~Nair, N.~Rathod, and K.~Yamada.
\newblock Unsupervised analysis for decipherment problems.
\newblock In \emph{Proceedings of the COLING/ACL}, pages 499--506. Association
  for Computational Linguistics, 2006.

\bibitem[Knight et~al.(2011)Knight, Megyesi, and Schaefer]{Knight11}
K.~Knight, B.~Megyesi, and C.~Schaefer.
\newblock The {Copiale} cipher.
\newblock In \emph{Proceedings of the 4th Workshop on Building and Using
  Comparable Corpora: Comparable Corpora and the Web}. Association for
  Computational Linguistics, 2011.

\bibitem[Kozielski et~al.(2014)Kozielski, Nuhn, Doetsch, and Ney]{Kozielski14}
M.~Kozielski, M.~Nuhn, P.~Doetsch, and H.~Ney.
\newblock Towards unsupervised learning for handwriting recognition.
\newblock In \emph{Frontiers in Handwriting Recognition (ICFHR), 2014 14th
  International Conference on}, pages 549--554. IEEE, 2014.

\bibitem[Krizhevsky et~al.(2012)Krizhevsky, Sutskever, and
  Hinton]{Krizhevsky12}
A.~Krizhevsky, I.~Sutskever, and G.~E. Hinton.
\newblock {ImageNet} classification with deep convolutional neural networks.
\newblock In \emph{Proc. {NIPS}}, pages 1106--1114, 2012.

\bibitem[Lample et~al.(2017)Lample, Denoyer, and Ranzato]{Lample17}
G.~Lample, L.~Denoyer, and M.~Ranzato.
\newblock Unsupervised machine translation using monolingual corpora only.
\newblock In \emph{Proc. {ICLR}}, 2017.

\bibitem[LeCun et~al.(1989)LeCun, Boser, Denker, Henderson, Howard, Hubbard,
  and Jackel]{Lecun89}
Y.~LeCun, B.~Boser, J.~S. Denker, D.~Henderson, R.~E. Howard, W.~Hubbard, and
  L.~D. Jackel.
\newblock Backpropagation applied to handwritten zip code recognition.
\newblock \emph{Neural Computation}, 1\penalty0 (4):\penalty0 541--551, 1989.

\bibitem[Lee and Osindero(2016)]{Lee16}
C.~Lee and S.~Osindero.
\newblock Recursive recurrent nets with attention modeling for {OCR} in the
  wild.
\newblock In \emph{Proc. CVPR}, 2016.

\bibitem[Lee et~al.(2014)Lee, Bhardwaj, Di, Jagadeesh, and Piramuthu]{Lee14}
C.~Lee, A.~Bhardwaj, W.~Di, V.~Jagadeesh, and R.~Piramuthu.
\newblock Region-based discriminative feature pooling for scene text
  recognition.
\newblock In \emph{Proc. CVPR}, 2014.

\bibitem[Lee(2002)]{Lee02}
D.-S. Lee.
\newblock Substitution deciphering based on {HMM}s with applications to
  compressed document processing.
\newblock \emph{{PAMI}}, \penalty0 (12):\penalty0 1661--1666, 2002.

\bibitem[Levenshtein(1966)]{Levenshtein66}
V.~Levenshtein.
\newblock Binary codes capable of correcting deletions, insertions and
  reversals.
\newblock In \emph{Soviet Physics Doklady}, volume~10, page 707, 1966.

\bibitem[Li and Wand(2016)]{Li16}
C.~Li and M.~Wand.
\newblock Precomputed real-time texture synthesis with markovian generative
  adversarial networks.
\newblock In \emph{Proc. {ECCV}}, pages 702--716. Springer, 2016.

\bibitem[Liu et~al.(2017{\natexlab{a}})Liu, Breuel, and Kautz]{Liu17ii}
M.-Y. Liu, T.~Breuel, and J.~Kautz.
\newblock Unsupervised image-to-image translation networks.
\newblock In \emph{Proc. {NIPS}}, pages 700--708, 2017{\natexlab{a}}.

\bibitem[Liu et~al.(2017{\natexlab{b}})Liu, Chen, and Deng]{liu17}
Y.~Liu, J.~Chen, and L.~Deng.
\newblock Unsupervised sequence classification using sequential output
  statistics.
\newblock In \emph{Proc. {NIPS}}, pages 3550--3559, 2017{\natexlab{b}}.

\bibitem[Long et~al.(2015)Long, Shelhamer, and Darrell]{Long15}
J.~Long, E.~Shelhamer, and T.~Darrell.
\newblock Fully convolutional networks for semantic segmentation.
\newblock In \emph{Proc. {CVPR}}, 2015.

\bibitem[Maas et~al.(2013)Maas, Hannun, and Ng]{Maas13}
A.~L. Maas, A.~Y. Hannun, and A.~Y. Ng.
\newblock Rectifier nonlinearities improve neural network acoustic models.
\newblock In \emph{Proc. {ICML}}, volume~30, page~3, 2013.

\bibitem[Mao et~al.(2017)Mao, Li, Xie, Lau, Wang, and Smolley]{Mao17}
X.~Mao, Q.~Li, H.~Xie, R.~Y. Lau, Z.~Wang, and S.~P. Smolley.
\newblock Least squares generative adversarial networks.
\newblock In \emph{Proc. {ICCV}}, pages 2813--2821. IEEE, 2017.

\bibitem[Mishra et~al.(2012{\natexlab{a}})Mishra, Alahari, and
  Jawahar]{Mishra12}
A.~Mishra, K.~Alahari, and C.~Jawahar.
\newblock Scene text recognition using higher order language priors.
\newblock \emph{Proc. {BMVC}}, 2012{\natexlab{a}}.

\bibitem[Mishra et~al.(2012{\natexlab{b}})Mishra, Alahari, and
  Jawahar]{Mishra12a}
A.~Mishra, K.~Alahari, and C.~Jawahar.
\newblock Top-down and bottom-up cues for scene text recognition.
\newblock In \emph{Proc. CVPR}, 2012{\natexlab{b}}.

\bibitem[Nagy(1986)]{Nagy86}
G.~Nagy.
\newblock Efficient algorithms to decode substitution ciphers with applications
  to {OCR}.
\newblock In \emph{Proc. {ICPR}}, pages 352--355, 1986.

\bibitem[Netzer et~al.(2011)Netzer, Wang, Coates, Bissacco, Wu, and
  Ng]{netzer2011reading}
Y.~Netzer, T.~Wang, A.~Coates, A.~Bissacco, B.~Wu, and A.~Y. Ng.
\newblock Reading digits in natural images with unsupervised feature learning.
\newblock In \emph{NIPS DLW}, volume 2011, 2011.

\bibitem[Neumann and Matas(2012)]{Neumann12}
L.~Neumann and J.~Matas.
\newblock Real-time scene text localization and recognition.
\newblock In \emph{Proc. {CVPR}}, volume~3, pages 1187--1190. IEEE, 2012.

\bibitem[Novikova et~al.(2012)Novikova, Barinova, Kohli, and
  Lempitsky]{Novikova12}
T.~Novikova, O.~Barinova, P.~Kohli, and V.~Lempitsky.
\newblock Large-lexicon attribute-consistent text recognition in natural
  images.
\newblock In \emph{Proc. {ECCV}}, pages 752--765. Springer, 2012.

\bibitem[Nuhn and Ney(2013)]{Nuhn13}
M.~Nuhn and H.~Ney.
\newblock Decipherment complexity in 1: 1 substitution ciphers.
\newblock In \emph{Proceedings of the 51st Annual Meeting of the Association
  for Computational Linguistics}, volume~1, pages 615--621, 2013.

\bibitem[Parkinson(1805)]{Parkinson1805}
J.~Parkinson.
\newblock \emph{Observations on the Nature and Cure of Gout: On Nodes of the
  Joints; and on the Influence of Certain Articles of Diet, in Gout,
  Rheumatism, and Gravel}.
\newblock Symonds, 1805.

\bibitem[Peleg and Rosenfeld(1979)]{Peleg79}
S.~Peleg and A.~Rosenfeld.
\newblock Breaking substitution ciphers using a relaxation algorithm.
\newblock \emph{Communications of the ACM}, 22\penalty0 (11):\penalty0
  598--605, 1979.

\bibitem[Poznanski and Wolf(2016)]{Poznanski16}
A.~Poznanski and L.~Wolf.
\newblock {CNN-N-Gram} for handwriting word recognition.
\newblock In \emph{Proc. CVPR}, 2016.

\bibitem[Ravi and Knight(2008)]{Ravi08}
S.~Ravi and K.~Knight.
\newblock Attacking decipherment problems optimally with low-order n-gram
  models.
\newblock In \emph{proceedings of the conference on Empirical Methods in
  Natural Language Processing}, pages 812--819. Association for Computational
  Linguistics, 2008.

\bibitem[Rodriguez-Serrano et~al.(2015)Rodriguez-Serrano, Gordo, and
  Perronnin]{Rodriguez15}
J.~A. Rodriguez-Serrano, A.~Gordo, and F.~Perronnin.
\newblock Label embedding: A frugal baseline for text recognition.
\newblock \emph{International Journal of Computer Vision}, 113\penalty0
  (3):\penalty0 193--207, 2015.

\bibitem[Shi et~al.(2015)Shi, Bai, and Yao]{ShiBY15}
B.~Shi, X.~Bai, and C.~Yao.
\newblock An end-to-end trainable neural network for image-based sequence
  recognition and its application to scene text recognition.
\newblock \emph{ArXiv e-prints}, 2015.

\bibitem[Shi et~al.(2016)Shi, Wang, Lv, Yao, and Bai]{Shi16}
B.~Shi, X.~Wang, P.~Lv, C.~Yao, and X.~Bai.
\newblock Robust scene text recognition with automatic rectification.
\newblock In \emph{Proc. CVPR}, 2016.

\bibitem[Shi et~al.(2013)Shi, Wang, Xiao, Zhang, Gao, and Zhang]{Shi13}
C.~Shi, C.~Wang, B.~Xiao, Y.~Zhang, S.~Gao, and Z.~Zhang.
\newblock Scene text recognition using part-based tree-structured character
  detection.
\newblock In \emph{Proc. CVPR}, 2013.

\bibitem[Smith(2007)]{Smith07}
R.~Smith.
\newblock An overview of the {Tesseract} {OCR} engine.
\newblock In \emph{Document Analysis and Recognition, 2007. ICDAR 2007. Ninth
  International Conference on}, volume~2, pages 629--633. IEEE, 2007.

\bibitem[Snyder et~al.(2010)Snyder, Barzilay, and Knight]{Snyder10}
B.~Snyder, R.~Barzilay, and K.~Knight.
\newblock A statistical model for lost language decipherment.
\newblock In \emph{Proceedings of the 48th Annual Meeting of the Association
  for Computational Linguistics}, pages 1048--1057. Association for
  Computational Linguistics, 2010.

\bibitem[Srivastava et~al.(2015)Srivastava, Mansimov, and
  Salakhudinov]{Srivastava15}
N.~Srivastava, E.~Mansimov, and R.~Salakhudinov.
\newblock Unsupervised learning of video representations using lstms.
\newblock In \emph{Proc. {ICML}}, 2015.

\bibitem[Su and Lu(2014)]{Su14}
B.~Su and S.~Lu.
\newblock Accurate scene text recognition based on recurrent neural network.
\newblock In \emph{Proc. ACCV}, 2014.

\bibitem[Sutskever et~al.(2014)Sutskever, Vinyals, and Le]{Sutskever14}
I.~Sutskever, O.~Vinyals, and Q.~V. Le.
\newblock Sequence to sequence learning with neural networks.
\newblock In \emph{Proc. {NIPS}}, pages 3104--3112, 2014.

\bibitem[Sutskever et~al.(2016)Sutskever, Jozefowicz, Gregor, Rezende,
  Lillicrap, and Vinyals]{sutskever16}
I.~Sutskever, R.~Jozefowicz, K.~Gregor, D.~Rezende, T.~Lillicrap, and
  O.~Vinyals.
\newblock Towards principled unsupervised learning.
\newblock In \emph{ICLR workshop}, 2016.

\bibitem[{Tesseract OCR}(1985 -- 2018)]{tesseract}
{Tesseract OCR}.
\newblock https://github.com/tesseract-ocr/, 1985 -- 2018.

\bibitem[Tieleman and Hinton(2012)]{Tieleman12}
T.~Tieleman and G.~Hinton.
\newblock Lecture 6.5-rmsprop: Divide the gradient by a running average of its
  recent magnitude.
\newblock \emph{COURSERA: Neural networks for machine learning}, 4\penalty0
  (2):\penalty0 26--31, 2012.

\bibitem[Wang and Belongie(2010)]{Wang10b}
K.~Wang and S.~Belongie.
\newblock Word spotting in the wild.
\newblock In \emph{Proc. {ECCV}}. Springer, 2010.

\bibitem[Wang et~al.(2011)Wang, Babenko, and Belongie]{Wang11}
K.~Wang, B.~Babenko, and S.~Belongie.
\newblock End-to-end scene text recognition.
\newblock In \emph{Proc. {ICCV}}, pages 1457--1464. IEEE, 2011.

\bibitem[Wang et~al.(2012)Wang, Wu, Coates, and Ng]{Wang12}
T.~Wang, D.~J. Wu, A.~Coates, and A.~Y. Ng.
\newblock End-to-end text recognition with convolutional neural networks.
\newblock In \emph{Proc. {ICPR}}, pages 3304--3308. IEEE, 2012.

\bibitem[Yao et~al.(2014)Yao, Bai, Shi, and Liu]{Yao14}
C.~Yao, X.~Bai, B.~Shi, and W.~Liu.
\newblock Strokelets: A learned multi-scale representation for scene text
  recognition.
\newblock In \emph{Proc. CVPR}, 2014.

\bibitem[Zhu et~al.(2017)Zhu, Park, Isola, and Efros]{Zhu17}
J.-Y. Zhu, T.~Park, P.~Isola, and A.~A. Efros.
\newblock Unpaired image-to-image translation using cycle-consistent
  adversarial networks.
\newblock In \emph{Proc. {ICCV}}, 2017.

\end{thebibliography}
\clearpage

\appendix
\section{Visualising Real Book Recognition}\label{sec:appendix}
\vspace{3mm}
In the following pages, we show predictions of our method on a few samples
from the \emph{test} set of the \emph{real} book dataset.
Pages are first segmented into lines and then fed into our model for recognition (see \cref{fig:real});
full page images are shown for visual presentation only.
We use the improved \emph{skip-RNN} recognition network described in~\cref{sec:real}.
The ``ground-truth'' is not perfect, as it itself is output from Google's OCR engine.
$\blacksquare$ denotes the $\texttt{<UNK>}$ character class.
We note the excellent recognition accuracy of our method,
which is trained without any paired/labelled training examples.

\clearpage
\newgeometry{right=35mm,bottom=5mm}
\pagestyle{plain}
\begin{landscape}
    \resizebox{!}{15cm}{\includegraphics[width=\textwidth]{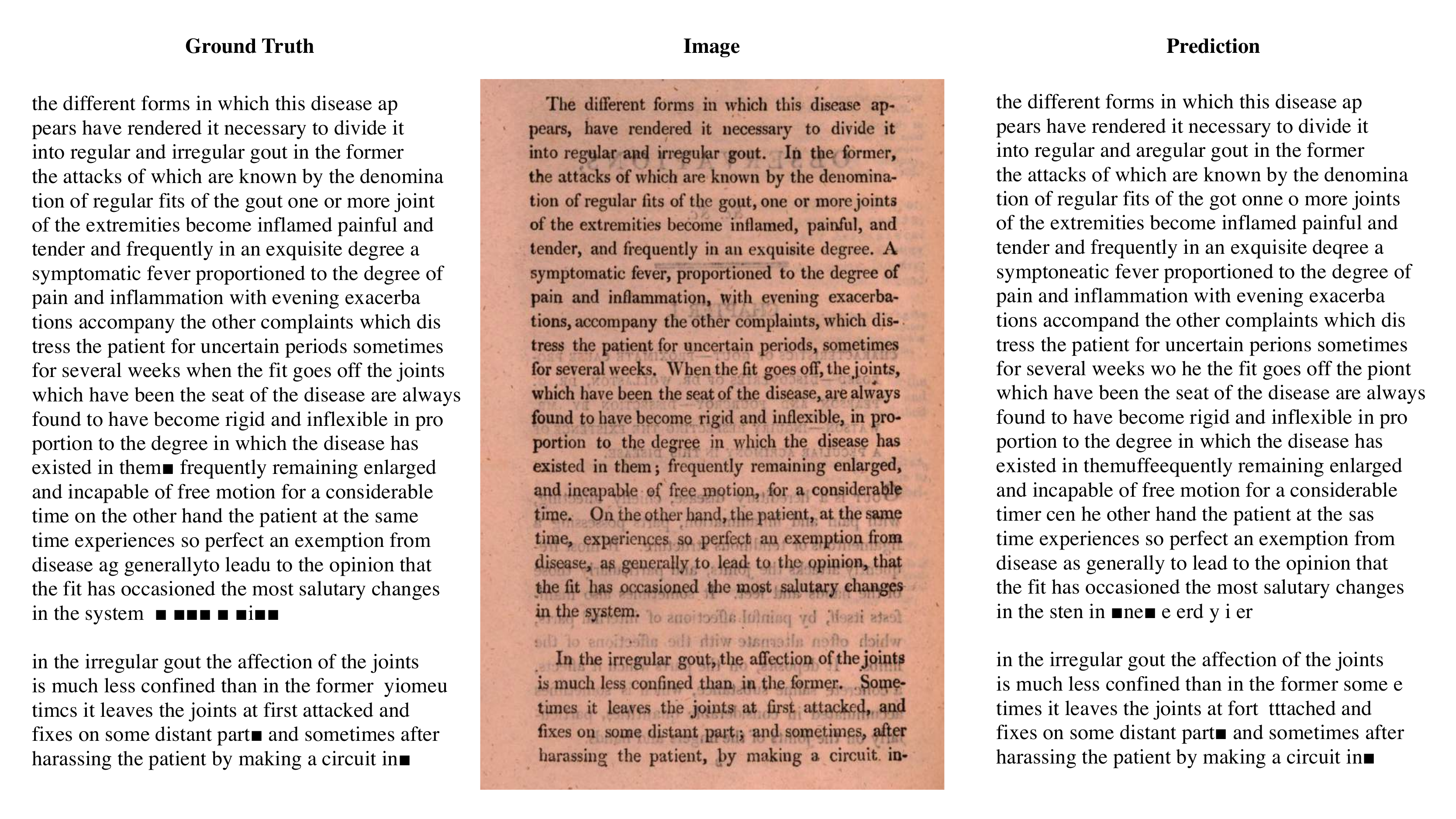}}\clearpage
    \resizebox{!}{15cm}{\includegraphics[width=\textwidth]{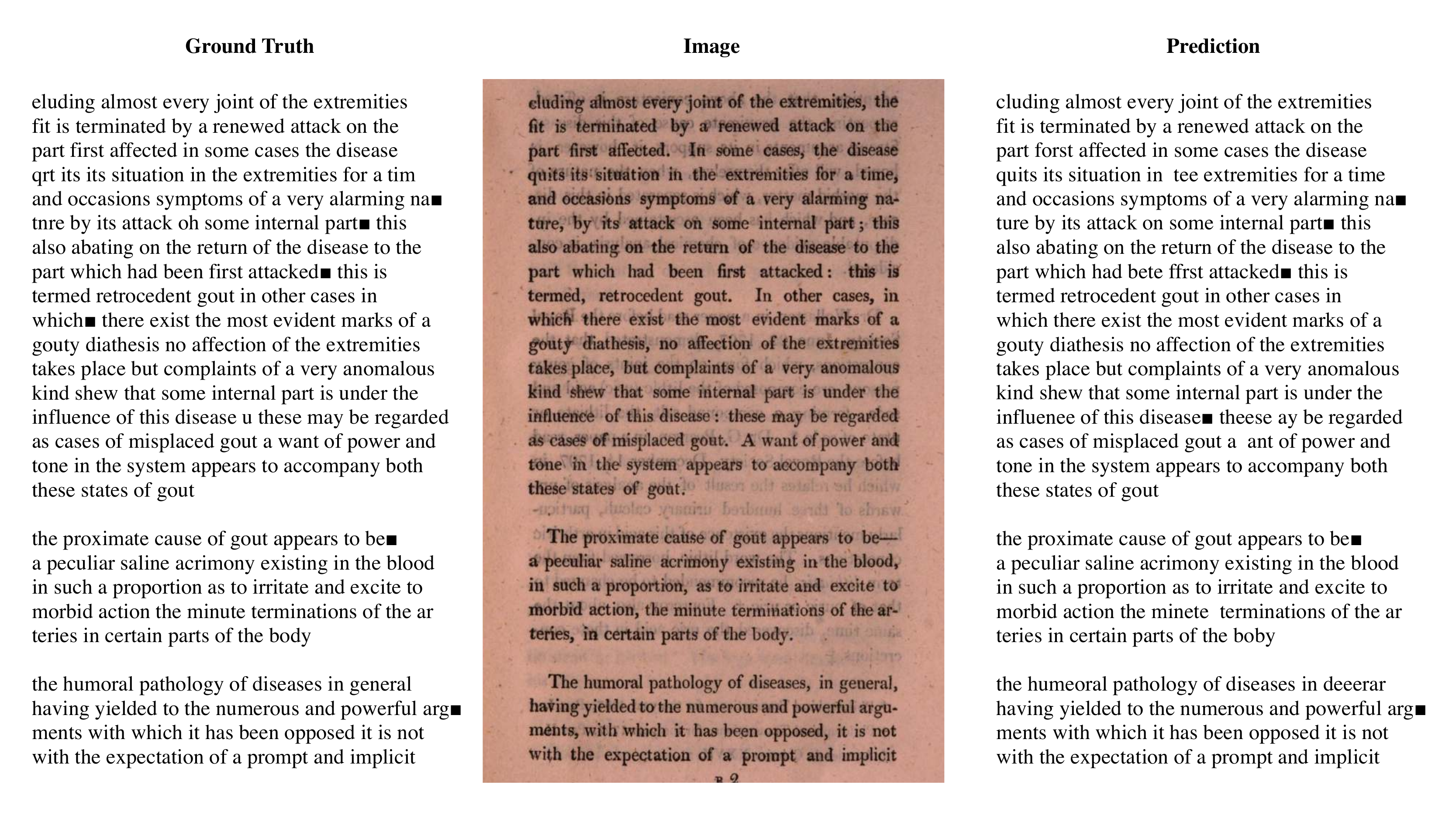}}\clearpage
    \resizebox{!}{15cm}{\includegraphics[width=\textwidth]{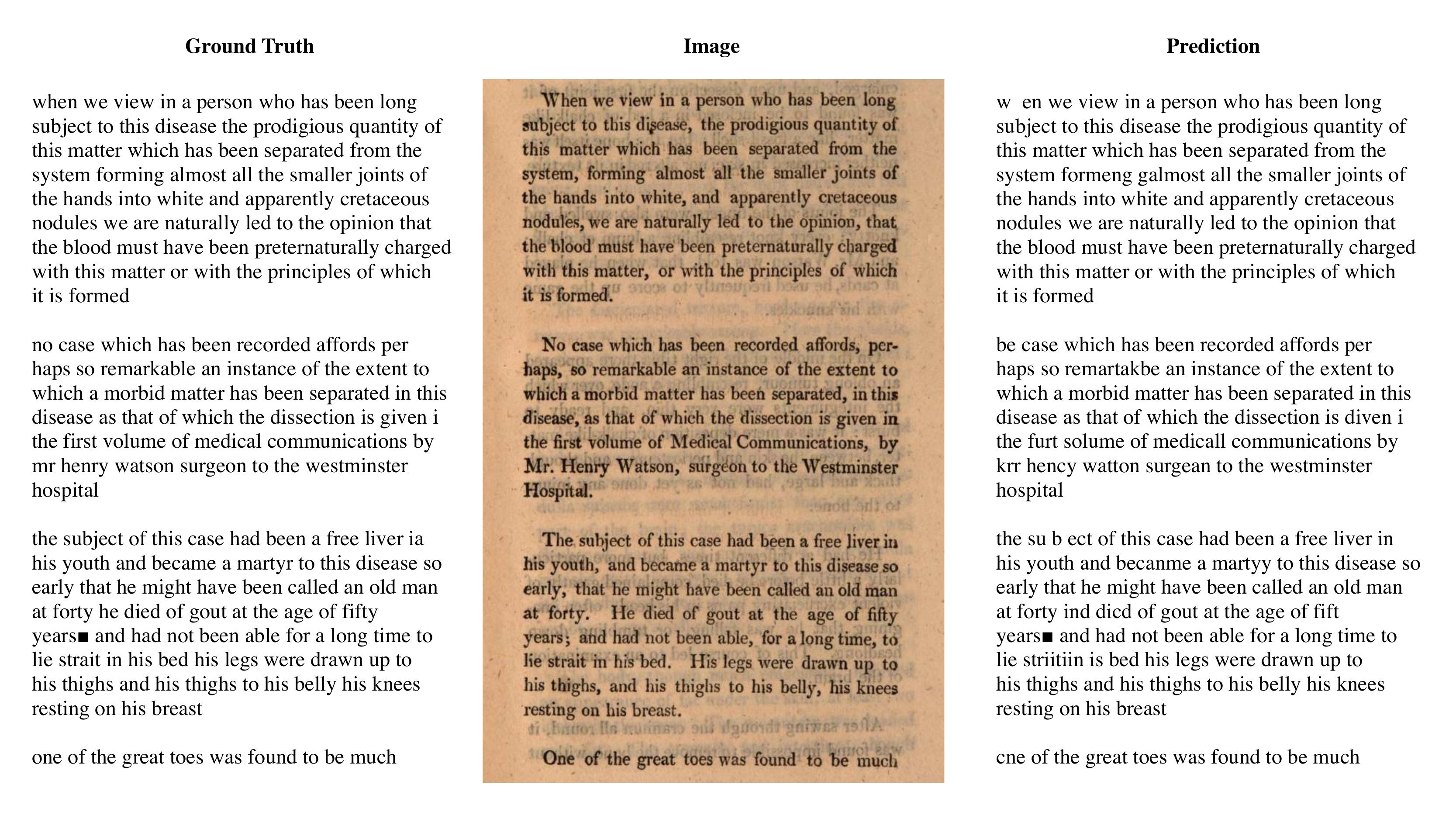}}\clearpage
\end{landscape}
\clearpage

\end{document}